\documentclass{article}

\PassOptionsToPackage{numbers, compress}{natbib}
\usepackage[preprint]{neurips_2026}
\usepackage{amsmath}
\usepackage{amssymb}
\usepackage{mathtools}
\usepackage{amsthm}
\usepackage{multirow}
\usepackage{tabularx}
\usepackage{bbding}
\usepackage{enumitem}

\usepackage[utf8]{inputenc} 
\usepackage[T1]{fontenc}    
\usepackage{float}          
\usepackage{wrapfig}        
\usepackage{capt-of}        
\usepackage{hyperref}       
\usepackage{url}            
\usepackage{booktabs}       
\usepackage{amsfonts}       
\usepackage{nicefrac}       
\usepackage{microtype}      
\usepackage{xcolor}         

\title{OrthoPhys: Physically Plausible Video Generation with Orthogonal-View Geometry Guidance}

\author{%
  \normalfont
  {\bfseries
  Cong Wang\textsuperscript{1,2,3 *},
  Hanxin Zhu\textsuperscript{4 *},
  Xiao Tang\textsuperscript{5},
  Jiayi Luo\textsuperscript{3,6}}\\
  {\bfseries
  Xin Jin\textsuperscript{3,7},
  Long Chen\textsuperscript{1 $\dagger$},
  Zhibo Chen\textsuperscript{3,4 $\dagger$}}\\
  \textsuperscript{1}the State Key Laboratory of Multimodal Artificial Intelligence Systems,\\ Institute of Automation, Chinese Academy of Sciences \\
  \textsuperscript{2}the School of Artificial Intelligence, University of Chinese Academy of Sciences \\
  \textsuperscript{3}Zhongguancun Academy \\
  \textsuperscript{4}School of Information Science and Technology, University of Science and Technology of China \\
  \textsuperscript{5}College of Automotive and Energy Engineering, Tongji University \\
  \textsuperscript{6}SKLCCSE, School of Computer Science and Engineering, Beihang University \\
  \textsuperscript{7}Eastern Institute of Technology \\
  \textsuperscript{*}Equal contribution 
  \textsuperscript{$\dagger$}Corresponding author
}

\begin{document}

\maketitle

\begin{abstract}
  Recent progress in video generation has led to substantial improvements in visual fidelity, yet ensuring physically consistent motion remains a fundamental challenge. Intuitively, this limitation can be attributed to the fact that real-world object motion unfolds in three-dimensional space, while video observations provide only partial, view-dependent projections of such dynamics. To address these issues, we propose \textbf{OrthoPhys}, a two-stage framework that leverages orthogonal-view geometry guidance to enforce physical plausibility. Instead of directly generating unstructured 2D videos, our first stage generates synchronized, four-view orthogonal videos of the foreground dynamics. By incorporating a geometry-enhanced attention mechanism across these orthogonal views, this stage effectively enforces 3D spatial coherence and implicitly grounds the motion in physical attributes. In the second stage, these physically consistent orthogonal foregrounds serve as rigid guidance to synthesize the final complete video, seamlessly learning the interaction between foreground dynamics and the background context. To support this orthogonal-view training paradigm, we construct \textbf{PhysMV}, a dataset containing 40K scenes, each consisting of four orthogonal viewpoints, resulting in a total of 160K video sequences. Extensive experiments demonstrate that OrthoPhys significantly improves physical realism and spatial-temporal coherence over existing video generation methods. Project page: \url{https://anonymous.4open.science/w/Phys4D/}.
\end{abstract}

\section{Introduction}
Recent advances in video generation have significantly expanded the capability of visual synthesis models, enabling the generation of high-fidelity and temporally coherent videos conditioned on multimodal inputs such as text and images~\cite{blattmann2023stable, zheng2024open, kong2024hunyuanvideo, yangcogvideox, wan2025wan}.
In particular, text- and image-conditioned video generation (TI2V) has emerged as a powerful paradigm for synthesizing dynamic visual content that aligns semantic intent with visual appearance.
As perceptual quality improves, these models are increasingly seen as a foundation for interactive content creation, simulation-based reasoning, and embodied AI in dynamic visual environments.

Despite this progress, current video generation models still struggle to produce physically plausible and spatially consistent motion~\cite{bansal2024videophy, meng2024towards}.
While visually convincing at the pixel level, generated videos
\begin{wrapfigure}{r}{0.50\linewidth}
    \centering
    \includegraphics[width=\linewidth]{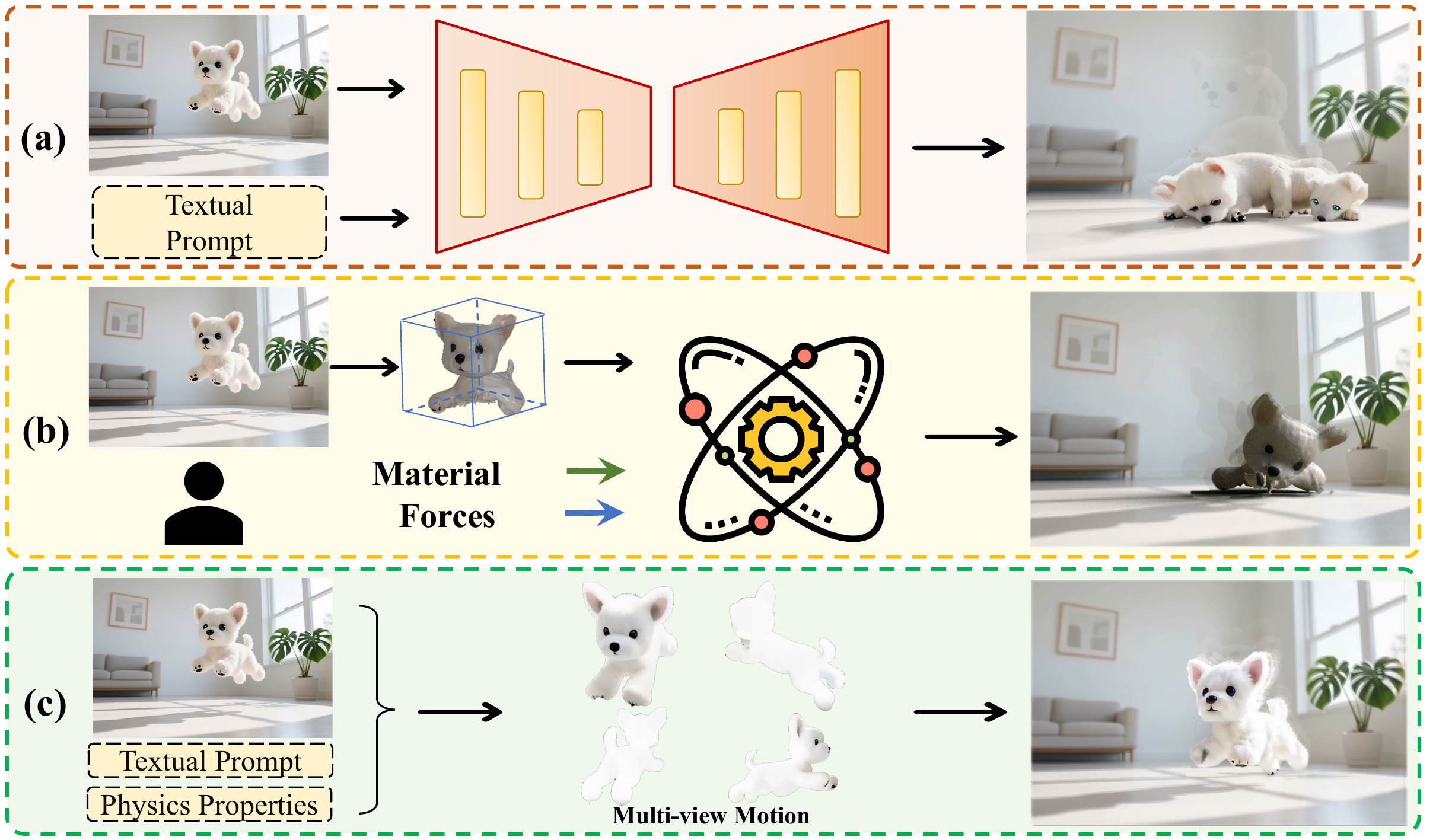}
    \caption{\textbf{Comparison with prior paradigms}. (a) Data-driven generative models. (b) Physics-engine-based methods. (c) Our physics-aware two-stage framework utilizing foreground multi-view videos.}
    \label{fig:teaser}
    \vspace{-0.35cm}
\end{wrapfigure} often violate basic physical principles, exhibiting implausible accelerations, inconsistent object interactions, or motion patterns that are insensitive to intrinsic object properties such as mass or elasticity, as shown in Fig.~\ref{fig:teaser}(a).
These issues become particularly evident in background-aware synthesis, where realistic interactions between moving foreground objects and the surrounding scene are essential.



We attribute these limitations to a mismatch between pixel-space motion modeling and physically constrained motion. Existing video generators typically learn dynamics as appearance transformations 
across frames, yet physically plausible motion and deformation are governed by geometry and material properties and should remain consistent across viewpoints. Explicit 3D representations and physics-based simulation~\cite{xie2024physgaussian, lin2025phys4dgen, liu2024physics3d} can produce valid trajectories, but they often rely on multi-stage pipelines that are difficult to scale and hard to integrate into end-to-end training. Instead, we consider a setting that directly outputs synchronized orthogonal-view foreground motion conditioned on physical attributes, and evaluates physical realism through both input-consistency with the specified physical properties and cross-view consistency, without explicitly outputting 3D structure or performing explicit 3D modeling.

In this work, we take a foreground-first view of physically grounded motion: we generate synchronized orthogonal-view foreground dynamics under explicit physical attributes, and then use them to guide final video synthesis. Building on this idea, we propose \textbf{OrthoPhys}, a two-stage framework that enables orthogonal-view geometry-guided motion modeling for physically plausible video generation without explicitly reconstructing 3D geometry. OrthoPhys adopts a two-stage design.
In the first stage, the \textbf{Phys4View} module generates orthogonal-view foreground motion conditioned on a single image, a textual prompt, and explicit physical attributes.
By injecting physical parameters directly into the attention mechanism and conditioning on control videos constructed from foreground masks and orthogonal-view priors, Phys4View incorporates physical and geometric cues into motion generation. We further introduce a geometry-aware cross-view attention module together with temporal attention to enhance cross-view alignment and spatiotemporal coherence, enabling simultaneous generation of synchronized four-view videos.
In the second stage, the \textbf{VideoSyn} module synthesizes background-aware image-to-video results by leveraging the generated orthogonal-view motion as guidance, allowing the model to learn data-driven foreground--background interactions such as contact, occlusion, and context-consistent motion.
To support training of the orthogonal-view video generation model, we employ a physics engine together with 3D-GS object representations to construct a physics-driven multi-view dataset, \textbf{PhysMV}, which comprises 10K 3D objects and 40K orthogonal video sets, totaling 160K videos.


Our contributions can be summarized as follows:
\begin{itemize}[left=0pt]
    \item We introduce \textbf{OrthoPhys}, a feed-forward framework for physically plausible video generation with orthogonal-view geometry guidance, conditioned on images, text, and explicit physical attributes.
    \item We design the \textbf{Phys4View} module, which integrates physics-aware attention with spatiotemporal and cross-view modeling to generate temporally coherent and geometrically consistent orthogonal-view motion.
    \item We develop the \textbf{VideoSyn} module for background-aware image-to-video synthesis, where orthogonal-view motion guidance enables data-driven learning of realistic foreground--background interactions without explicit physical simulation at inference.
    \item We construct a large-scale physics-oriented multi-view video dataset using a physics engine. We further show that OrthoPhys, leveraging orthogonal-view foreground motion generation followed by motion-aware video synthesis, achieves superior physical realism, temporal coherence, and cross-view consistency compared to prior video generation methods.

\end{itemize}

\section{Related Works}
\subsection{Controllable Video Generation}
Video generation models trained on large-scale text–video paired datasets have demonstrated remarkable capabilities in synthesizing high-quality videos~\cite{ho2022video, blattmann2023stable, kong2024hunyuanvideo, yangcogvideox}. Previous studies have shown that pretrained models can be additionally guided by various control signals, including camera motion~\cite{fu20243dtrajmaster, he2024cameractrl}, point trajectories~\cite{geng2025motion, gu2025diffusion, burgert2025go}, and anchor-frame videos~\cite{chen2025stance, romero2025learning}, enabling more controllable video generation. Meanwhile, other research efforts have explored leveraging different modalities to guide motion-aware video synthesis. Some approaches~\cite{li2024generative, jin2025flovd} employ optical flow videos as guidance, using flow maps to characterize motion dynamics and subsequently synthesize RGB videos based on them. Other methods~\cite{liang2024flowvid, lv2024gpt4motion} use depth videos as conditioning inputs to guide the generation of temporally coherent video sequences. In addition, several text-driven approaches~\cite{xue2025phyt2v} attempt to infer more fine-grained descriptions of motion dynamics from textual instructions, thereby enabling more precise control over motion behaviors during video generation.
However, these methods generally lack the modeling of physical laws and often produce results that violate basic principles of physical plausibility. 
To address this limitation, OrthoPhys generates physics-aware orthogonal-view videos as motion representations and subsequently renders them into plausible video outputs.

\begin{figure*}
    \centering
    \includegraphics[width=1\linewidth]{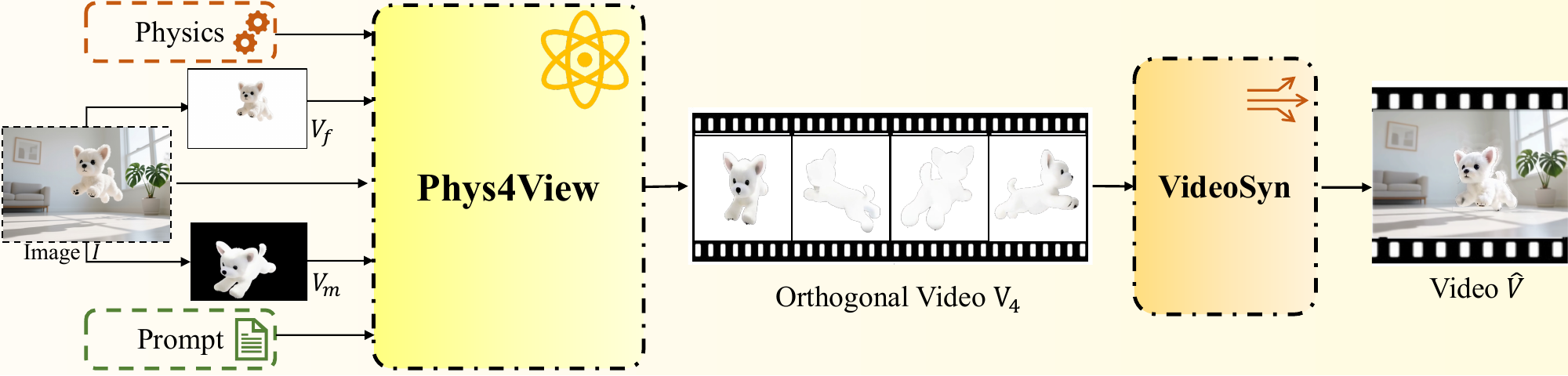}
    \caption{\textbf{Pipeline of OrthoPhys}. OrthoPhys generates physically plausible videos via a two-stage pipeline: 1) generating physics-aware orthogonal-view foreground videos, 2) generating a plausible video with the guidance of foreground motion.}
    \label{fig:framework}
    \vspace{-0.5cm}
\end{figure*}

\subsection{Physics-grounded Video Generation}
Physics engines typically use the 3D representation of foreground objects as input to simulate and generate the 3D state at each time step, based on defined material and motion properties~\cite{xie2024physgaussian, chen2025physgen3d}. Leveraging the renderable 3D-GS representation~\cite{kerbl20233d}, some methods~\cite{xie2024physgaussian} propose using the Material Point Method (MPM) solver to update the 3D Gaussian representation at each time step after force-induced motion. However, due to the limitations of MPM, it is only suitable for simulating elastic objects. To address the shortcomings of MPM solvers, some approaches~\cite{feng2024gaussian, gao2025fluidnexus} incorporate Position-Based Dynamics (PBD) solvers to simulate fluid motion. The use of physics engines requires additional definition of physical properties, and some methods~\cite{zhao2025physsplat, mao2025live} propose using MLLMs to estimate properties such as Young's modulus, Poisson's ratio, and density. Furthermore, other methods~\cite{huang2025dreamphysics, lin2025omniphysgs, zhang2024physdreamer} suggest utilizing the prior knowledge of pre-trained video generation models to iteratively optimize the material properties of objects through SDS loss~\cite{poole2022dreamfusion} after rendering the video.
\textcolor{black}{Although physics-engine-based methods ensure physical plausibility, they require manual definition of simulation conditions such as material properties and external forces and depend on high-quality explicit 3D representations, limiting their ability to automatically generate high-quality videos from a single image.}


\section{Methodology}\label{sec:method}
We propose \textbf{OrthoPhys}, a two-stage generation framework that first produces orthogonal-view videos with physically grounded motion and subsequently synthesizes visually realistic videos with foreground–background interactions, as shown in Fig.~\ref{fig:framework}. Its orthogonal-view generation module, \textbf{Phys4View}, is presented in Sec.~\ref{sec:phys4view1} and Sec.~\ref{sec:phys4view2}, followed by its video synthesis module, \textbf{VideoSyn}, in Sec.~\ref{sec:videosyn}.

\subsection{Physics Conditioning and Video Prior Fusion}\label{sec:phys4view1}

Phys4View aims to generate temporally coherent and geometrically consistent orthogonal-view videos conditioned on a single image $I$, a textual prompt $T$, and physical attributes $P$. Given an input image containing both foreground and background regions, we first extract a foreground object mask $M_f$ using a semantic segmentation model~\cite{ren2024grounded}. Based on $M_f$, we isolate the foreground image $I_f$ and construct a foreground-initialized video $V_f$ by replicating $I_f$ across the temporal dimension. To provide 3D structural priors for subsequent video generation, we further employ a pretrained image-to-3D generation model~\cite{xiang2025structured} to obtain a static multi-view video $V_m$, which serves as geometric guidance.

\noindent \textbf{Video prior conditioning.}
We condition video generation on two complementary video priors: a foreground-initialized video \( V_f \) and an orthogonal-view structural prior \( V_m \).
Both priors are encoded into latent features \( H_f \) and \( H_m \), and are adaptively fused via a lightweight gating mechanism.
Specifically, we predict a gating tensor from their concatenation and compute the fused representation as
\begin{equation}
    H_v = G \odot H_f + (1 - G) \odot H_m,
\end{equation}
where \( G = \mathrm{Proj}(H_f \oplus H_m) \) and \( \mathrm{Proj}(\cdot) \) is a lightweight 3D convolutional projector.
We then inject \( H_v \) into the intermediate hidden representation \( H_i \) by replacing a designated subset of channels:
\begin{equation}
    H_c = \mathrm{Inject}(H_i, H_v).
\end{equation}


\begin{figure*}
    \centering
    \includegraphics[width=1\linewidth]{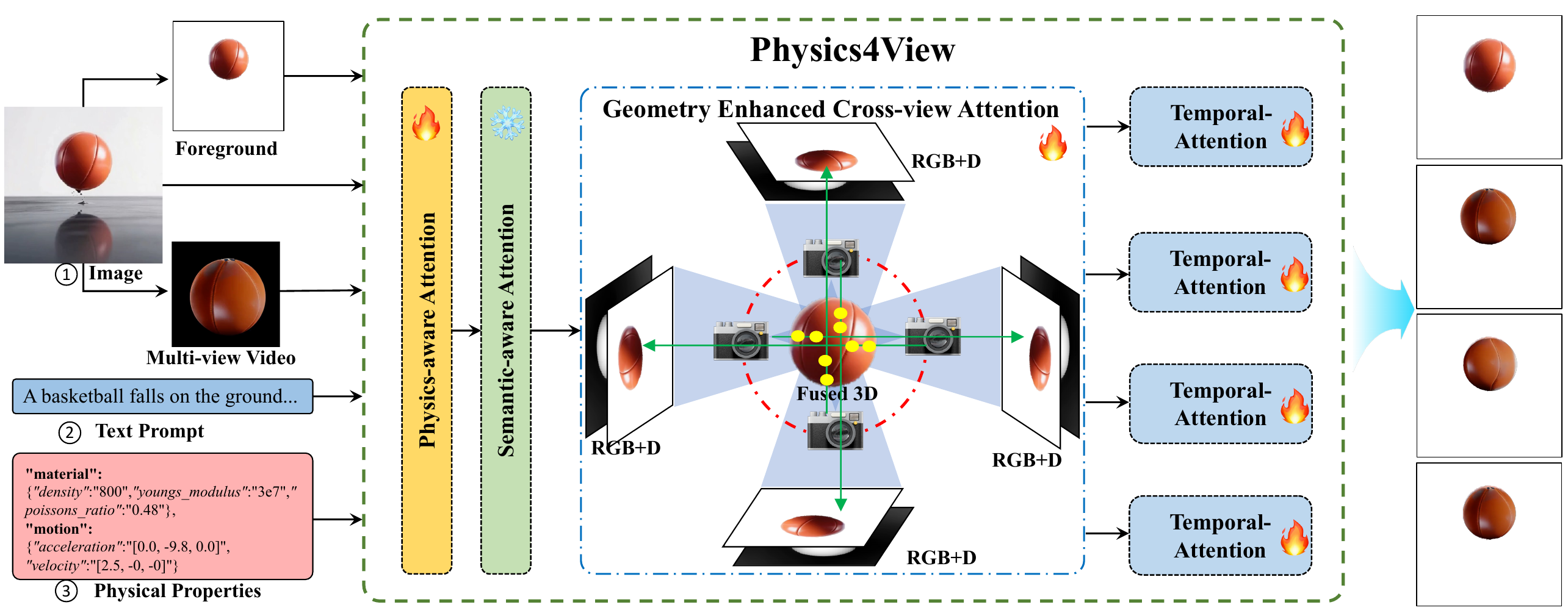}
    \vspace{-0.5cm}
    \caption{\textbf{Overview of Phys4View}. This stage of OrthoPhys incorporates physics-aware attention to model physical motion and introduces a geometry-enhanced cross-view attention module to generate spatially and temporally consistent orthogonal-view videos.}
    \label{fig:phys4view}
    \vspace{-0.5cm}
\end{figure*}

\noindent \textbf{Physical and semantic conditioning.}
We condition generation on object-level physical attributes, including motion cues (e.g., acceleration \( \mathbf{a} \) and velocity \( \mathbf{v} \)) and material properties (e.g., density \( \rho \), Young's modulus \( E \), and Poisson's ratio \( \nu \)).
To avoid semantic distortion from an expressive attribute encoder, we adopt a parameter-free tokenization scheme and obtain a unified physical conditioning token sequence \( C_{\text{phys}} \).

We incorporate \( C_{\text{phys}} \) into the generator via a physics-aware attention module that modulates the intermediate representation \( H_c \):
\begin{equation}
    H_{\text{phys}} = H_c + \alpha_{\text{phys}} \cdot \mathrm{Attn}_{\text{phys}}([H_c; C_{\text{phys}}]),
\end{equation}
where \( \alpha_{\text{phys}} \) is a learnable scaling factor.
For textual conditioning, we follow the standard text cross-attention used in the pretrained TI2V model~\cite{yangcogvideox}:
\begin{equation}
    H = H_{\text{phys}} + \alpha_t \cdot \mathrm{Attn}_t(H_{\text{phys}}; C_p),
\end{equation}
where \( C_p \) denotes the encoded prompt.


\subsection{Geometry-enhanced Cross-view and Temporal Attention}\label{sec:phys4view2}
To prove spatial and temporal consistency, we introduce a geometry-enhanced cross-view and temporal attention, as illustrated in Fig.~\ref{fig:phys4view}.

\noindent \textbf{Camera pose fusion.}
Before applying spatiotemporal attention, we inject camera pose information into the four-view latent features in a lightweight manner. Specifically, we encode the per-view camera pose into a learnable embedding, concatenate it with the corresponding latent along the channel dimension, and use a small convolutional block to fuse them, yielding pose-aware features for subsequent cross-view interaction.

\noindent\textbf{Geometry-enhanced cross-view attention.}
To encourage spatially coherent cross-view interactions while avoiding explicit 3D representation, we introduce a geometry-guided bias into the cross-view attention mechanism based on predicted depth.

Given the latent feature map \( H \), we predict a dense depth map at the same spatial resolution:
\begin{equation}
    D = \mathcal{P}_\theta(H),
\end{equation}
where \( \mathcal{P}_\theta \) denotes a learnable depth prediction head.
Each spatial location \( i \) in the latent grid, associated with coordinates \( (u_i, v_i) \),
is lifted into the camera coordinate system via
\begin{equation}
    \mathbf{x}_{i}^{\mathrm{cam}}
    =
    D_i \, \mathbf{K}^{-1}
    \begin{bmatrix}
    u_i \\
    v_i \\
    1
    \end{bmatrix},
\end{equation}
where \( \mathbf{K} \) is the camera intrinsic matrix.

Using the reconstructed 3D points in camera coordinates,
we compute the pairwise geometric distance between locations \( i \) and \( j \) as
\begin{equation}
    d_{ij} = \left\lVert \mathbf{x}_i^{\mathrm{cam}} - \mathbf{x}_j^{\mathrm{cam}} \right\rVert_2.
\end{equation}
This distance is converted into a geometry-based affinity weight:
\begin{equation}
    w_{ij,d} = \exp\!\left(-\frac{d_{ij}^2}{\tau}\right),
\end{equation}
where \( \tau \) controls the spatial sensitivity of the geometric prior.

Since depth prediction may be unreliable near occlusion boundaries or textureless regions,
we further introduce a depth-confidence term to downweight ambiguous correspondences.
Specifically, we define:
\begin{equation}
    w_{ij,\text{conf}} =
    \alpha + (1-\alpha)
    \left(
    1 - f_{\theta}\!\left(
    \frac{\|\nabla D_i\|_2 + \|\nabla D_j\|_2}{|D_i| + |D_j| + \epsilon}
    \right)
    \right),
\end{equation}
where \( \nabla D_i \) denotes the local spatial depth gradient at location \( i \),
\( f_{\theta}(\cdot) \) is a learnable confidence mapping function,
\( \epsilon \) is a small constant for numerical stability,
and \( \alpha \) provides a lower bound on the confidence weight.

The final geometry-aware affinity between locations \( i \) and \( j \) is given by:
\begin{equation}
    w_{ij} = w_{ij,d} \cdot w_{ij,\text{conf}}.
\end{equation}

We incorporate this geometric prior into cross-view attention by augmenting the attention logits as:
\begin{equation}
    s_{ij}^{\text{geo}} = \frac{Q_i \cdot K_j}{\sqrt{d}} + \log(w_{ij}),
\end{equation}
where \( Q_i \) and \( K_j \) denote the query and key features at locations \( i \) and \( j \), respectively.
This formulation softly biases the attention mechanism toward spatially compatible and
geometrically plausible regions under the predicted depth, while suppressing unreliable interactions.










\noindent \textbf{Temporal attention.}
To further enhance temporal consistency in the generated videos, we introduce a lightweight temporal attention mechanism. Given the feature representation \( H \), we first partition it into four disjoint components along the channel dimension as $H = [H_1;\, H_2;\, H_3;\, H_4]$, where \( [\cdot;\cdot] \) denotes channel-wise concatenation.

For each component \( H_i \), we apply an independent temporal self-attention operation to model temporal dependencies:
\begin{equation}
    H'_i = H_i + \alpha_t \cdot \mathrm{Attn}(H_i),
\end{equation}
where \( \mathrm{Attn}(\cdot) \) denotes the temporal attention module and \( \alpha_t \) is a learnable scaling factor that controls the strength of temporal aggregation.


Finally, the temporally enhanced features from all groups are concatenated to form the final representation $H_{\text{final}} = [H'_1;\, H'_2;\, H'_3;\, H'_4]$.

\begin{wrapfigure}{r}{0.50\linewidth}
    \centering
    \includegraphics[width=1\linewidth]{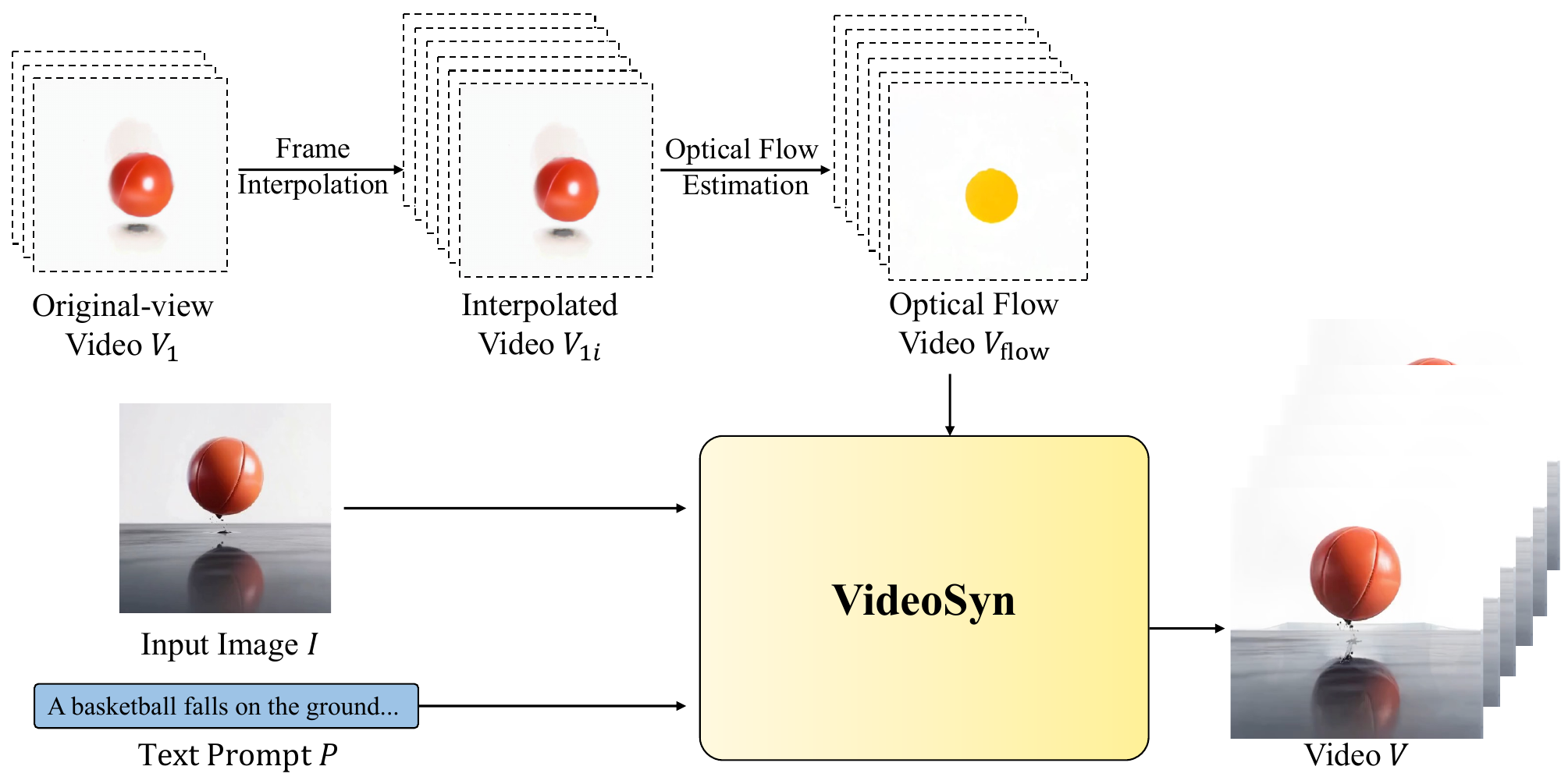}
    \caption{\textbf{Overview of VideoSyn}. The framework extracts foreground motion cues from a pre-generated physics-aware video and leverages them to guide the full video generation process.}
    \label{fig:videosyn}
    \vspace{-0.3cm}
\end{wrapfigure}

\subsection{Motion-guided Plausible Video Generation}
\label{sec:videosyn}

After obtaining a physics-aware four-view video \( V_4 \), our goal is to synthesize a visually coherent RGB video that captures realistic foreground–background interactions. To this end, we first extract the original-view video \( V_1 \) from \( V_4 \). We then apply a frame interpolation model~\cite{wu2024perception} to densify the temporal resolution by interpolating four intermediate frames between consecutive frames, resulting in an extended video sequence \( V_{1i} \).

To explicitly leverage the motion information encoded in \( V_{1i} \) as guidance for video generation, we adopt optical flow as a motion control signal. Specifically, we employ a pretrained optical flow estimation model~\cite{dong2024memflow} to compute the corresponding flow video \( V_{\mathrm{flow}} \) from \( V_{1i} \). The resulting optical flow captures fine-grained motion dynamics while remaining agnostic to appearance variations, making it a suitable representation for motion conditioning.

We then incorporate \( V_{\mathrm{flow}} \) as an explicit motion control condition in the video generation process:
\begin{equation}
    \tilde{V} = \mathrm{Attn}([H;\, H_{\mathrm{flow}};\, C_p]),
\end{equation}
where \( H_{\mathrm{flow}} \) is the encoded latent of the optical flow video \( V_{\mathrm{flow}} \). The attention module integrates appearance, motion, and semantic guidance for prediction.

The final video latent \( \hat{V} \) is obtained through the diffusion denoising process:
\begin{equation}
    \hat{V} = \mathcal{D}_{\theta}(\tilde{V}_t, t),
\end{equation}
where \( \tilde{V}_t \) denotes the noisy latent sampled at diffusion timestep \( t \).


\subsection{Training}
\paragraph{PhysMV dataset.}
To facilitate the training of our physics-aware, orthogonal-view video generation model, we construct PhysMV, a large-scale dataset comprising 40K distinct motion sequences, each captured from four orthogonal viewpoints, resulting in a total of 160K single-view videos.
To build up the dataset, we generate 10K foreground objects represented as 3D Gaussian Splatting (3D-GS) models using Trellis~\cite{xiang2025structured}. 
\textcolor{black}{Additional dataset details are provided in Appendix~\ref{sec:app_data}.}



\paragraph{Phys4View training.}
We train the proposed Phys4View model using a standard diffusion objective. Specifically, the base training loss is defined as
\begin{equation}
\label{eq:base-loss}
\mathcal{L}_{\mathrm{base}}
=
\mathbb{E}_{\tilde{F}, t}
\Bigl[
w_1(t)
\bigl\|
\mathcal{D}_\theta(\tilde{F}_t, t)
-
\bar{F}
\bigr\|_2^2
\Bigr],
\end{equation}
where \( \tilde{F}_t \) denotes the noisy latent at diffusion timestep \( t \), \( \bar{F} \) is the target latent, and \( w_1(t) \) is a timestep-dependent weighting function.

To provide geometric supervision, we additionally impose a depth consistency loss using relative depth estimates. Given the predicted depth map \( D \) and the estimated relative depth \( D_r \), we define the depth supervision loss as
\begin{equation}
    \mathcal{L}_{\text{depth}}
    =
    \left\|
    \frac{\mathrm{Cov}(D, D_r)}
    {\sqrt{\mathrm{Var}(D)\,\mathrm{Var}(D_r)}}
    \right\|_1,
    \label{loss:depth}
\end{equation}
which measures the scale-invariant correlation between the predicted and reference depth maps. 

\begin{figure*}
    \centering
    \includegraphics[width=1\linewidth]{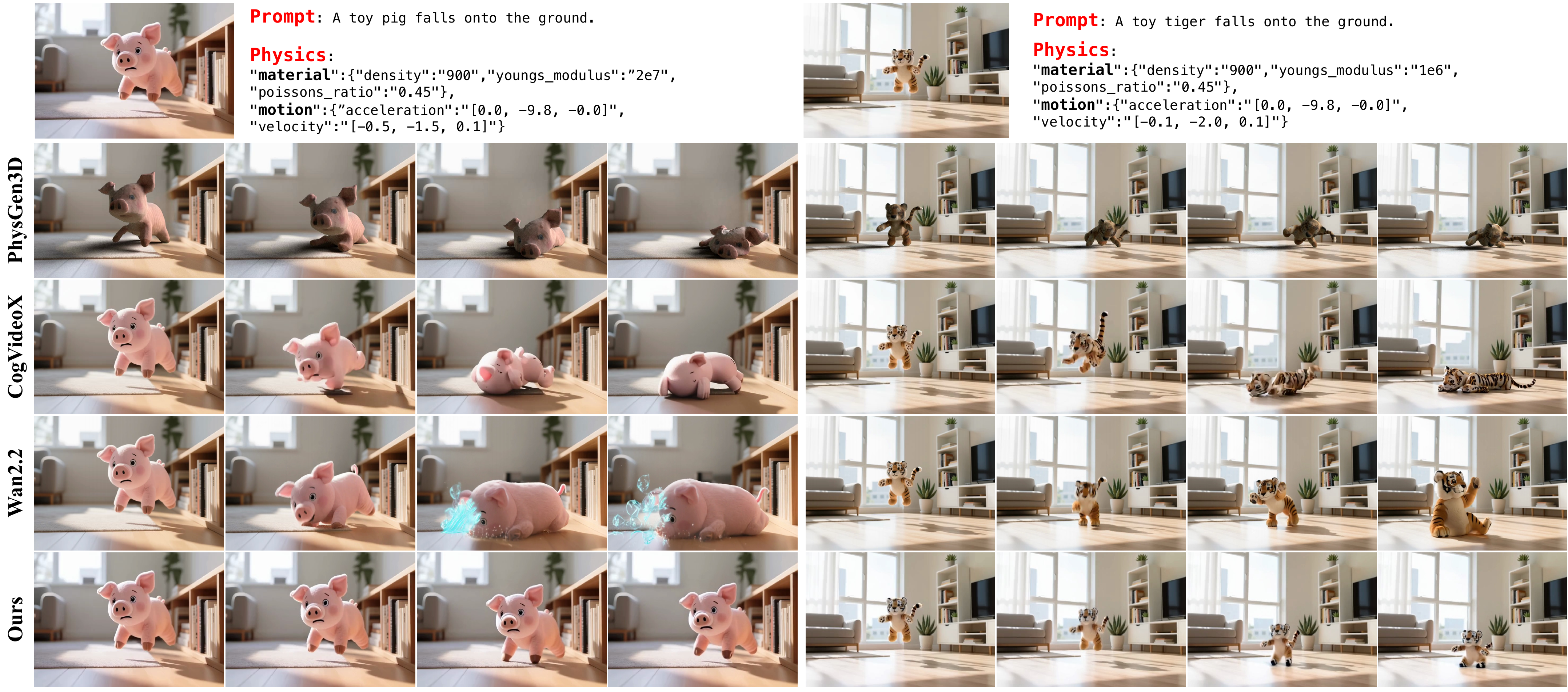}
    \vspace{-0.5cm}
    \caption{\textbf{Qualitative comparisons.} The top row shows the given image, text prompt and corresponding physics settings. For the definition of velocity and acceleration directions, the positive x-axis is defined as horizontal left, the positive y-axis as vertical upward, and the positive z-axis as perpendicular to the paper plane inward.}
    \label{fig:com}
    \vspace{-0.5cm}
\end{figure*}

\paragraph{VideoSyn training.}
To train VideoSyn, we adopt a diffusion-based image-to-video learning paradigm in which the model learns to generate a video conditioned on the motion guidance from the original view.
We preprocess the OpenVid~\cite{nan2024openvid} and WISA~\cite{wang2025wisa} datasets and train the VideoSyn model on both datasets.
During training, we minimize the mean squared error between the denoised prediction and the ground-truth frame under random diffusion steps:
\begin{equation}
    \label{eq:flowrender-loss}
    \mathcal{L}_{2}
    = 
    \mathbb{E}_{\tilde{{R}},\,t}
    \Bigl[w_2(t)
    \bigl\|
    \mathcal{D}_\theta(\tilde{R}_t, t)
    -
    \bar{R}
    \bigr\|_2^2
    \Bigr],
\end{equation}
where $\bar{R}$ is the ground-truth RGB video.
This training objective guides VideoSyn to synthesize temporally coherent and visually realistic video sequences that faithfully follow the motion guidance.

\begin{table*}[t]
    \centering
    \caption{\textbf{Quantitative comparisons on WorldScore and VBench.} \textbf{Bold}: Best. \underline{Underline}: Second Best.}
    \setlength{\tabcolsep}{3pt}
    \renewcommand{\arraystretch}{1.0} 
    \begin{tabularx}{\linewidth}{@{} lccccccc @{}}
    \toprule
    \multirow{2}{*}{\textbf{Method}} & \multicolumn{3}{c}{\textbf{WorldScore}} & \multicolumn{4}{c}{\textbf{VBench}} \\
    \cmidrule(lr){2-4} \cmidrule(lr){5-8}
    & Photo↑ & Motion↑ & 3D Consist↑ & Motion↑ & Subject↑ & Flickering↑ & Image↑  \\ \midrule
        CogVideoX-I2V-5B~\citep{yangcogvideox} & 64.72 & 67.39 & 74.43 & 0.993 & 0.886 & 0.989 & 0.636 \\
        Wan2.2-TI2V-5B~\citep{wan2025wan}   & 65.97 & 64.90 & 63.86  & 0.994 & 0.926 & 0.991 & \underline{0.653} \\
        Force Prompting~\cite{gillman2025force}  & 74.38 & 72.53 & 50.97 & 0.994 & 0.938 & 0.992 & 0.641 \\
        OmniPhysGS~\cite{lin2025omniphysgs}  & 58.67 & \underline{90.65} & 37.24  & 0.995 & 0.929 & 0.995 & 0.378 \\
        PhysGen~\cite{liu2024physgen}  & \underline{90.80} & 82.45 & 84.27  & 0.995 & \underline{0.956} & 0.994 & 0.648 \\
        PhysGen3D~\cite{chen2025physgen3d}  & 86.68 & 86.34 & \underline{86.89}  & \underline{0.996} & 0.910 & \textbf{0.998} & 0.605 \\
        \midrule
        \textbf{Ours}  & \textbf{94.09} & \textbf{93.62} &  \textbf{87.19} & \textbf{0.996} & \textbf{0.970} & \underline{0.996} & \textbf{0.655} \\
        \bottomrule
    \end{tabularx}%
    \label{tab:com}
    \vspace{-0.5cm}
\end{table*}

\section{Experiments}\label{sec:exp}
\subsection{Experimental Setups}\label{sec:exp_setup}

\paragraph{Implementation details.}
Our model is trained on 8 NVIDIA A100 GPUs, each with 80GB of memory, while all inference experiments are performed on a single A100 GPU.
To assess the effectiveness of the proposed approach, we construct a benchmark dataset consisting of 50 test samples. Each sample comprises a single RGB image \( I \) with a resolution of \(720 \times 480\), an associated textual prompt \( P \), and object-level physical attributes generated by GPT-4o. The benchmark set includes both rigid and deformable objects to evaluate the model’s generalization across diverse physical regimes. All input images are synthesized in a photorealistic style using a text-to-image generation model~\cite{wu2025qwen}.
For each input image, we obtain the foreground object mask \( M_f \) using Grounded-SAM~\cite{ren2024grounded}. Based on the extracted foreground, we further generate multi-view videos and corresponding 3D Gaussian Splatting (3D-GS) representations with Trellis~\cite{xiang2025structured}, which are used for downstream evaluation.
The comparison baselines are detailed in Appendix Sec.~\ref{sec:app_baseline_details}.

\paragraph{Metrics.}
We evaluate the quality of the generated videos using a combination of automatic metrics and VLM–based assessments. Specifically, WorldScore~\citep{duan2025worldscore} is used to quantify photometric consistency (Photo), multi-view geometric consistency (3D Consist), and motion smoothness (Motion). Complementarily, we employ VBench~\citep{huang2024vbench} to assess motion quality (Motion), subject consistency (Subject), temporal flickering (Flickering), and overall visual fidelity (Image).
To further measure adherence to the input prompts, following the evaluation protocol of PhysGen3D~\citep{chen2025physgen3d}, we utilize GPT-4o to score the generated videos along three aspects: physical realism, photorealism, and semantic alignment with the textual prompt.
We additionally conduct a fair blind user study with 59 participants under the same three criteria to complement the VLM-based assessment.

\subsection{Comparisons with State-Of-The-Art Methods}

\paragraph{Physics plausibility comparison.}
Quantitatively, as shown in Table~\ref{tab:com}, our method achieves superior motion coherence, consistently outperforming prior video generative models and showing comparable results with physical-engine-based methods across key dynamic metrics. Moreover, the generated videos exhibit higher visual quality than physical-engine-based methods. Regarding physical plausibility, as demonstrated in Table~\ref{tab:subjective}, our method surpasses all competing approaches under both GPT-4o evaluation and the blind user study, while simultaneously maintaining strong alignment with the input text and high semantic consistency.
Fig.~\ref{fig:com} shows two qualitative comparisons. Existing methods suffer from texture degradation, incorrect motion dynamics, or inconsistent object–background interactions, often leading to object replacement. In contrast, our method consistently generates physically plausible and coherent object motion. 
More comparisons can be found in Appendix Sec.~\ref{More visual comparisons with baselines}.


\begin{table}[t]
    \centering
    \caption{\textbf{GPT-4o and human evaluation results}. The left block reports GPT-4o scores, and the right block reports results from a fair blind user study with 59 participants. Both use three metrics: physical realism (Physical), photorealism (Photo), and semantic consistency (Semantic).}
    \vspace{-0.1cm}
    \setlength{\tabcolsep}{8pt}
    \renewcommand{\arraystretch}{1.0}
    \begin{tabularx}{\linewidth}{@{} l|ccc|ccc @{}}
    \toprule
        \multirow{2}{*}{Method} & \multicolumn{3}{c|}{GPT-4o} & \multicolumn{3}{c}{Human Study} \\
        \cmidrule(lr){2-4} \cmidrule(lr){5-7}
        & Physical↑ & Photo↑ & Semantic↑ & Physical↑ & Photo↑ & Semantic↑ \\ 
        \midrule
        CogVideoX-5B~\cite{yangcogvideox} &  0.612  &  0.858  & 0.617 & 2.80 & 3.14 & 3.14 \\
        Wan2.2-5B~\cite{wan2025wan} &  0.609  &  0.877  &  \underline{0.630}  & \underline{3.24} & \underline{3.41} & \underline{3.46} \\
        ForcePrompt~\cite{gillman2025force} &  0.615  &  0.875  &  0.611 & 2.25 & 2.41 & 2.74 \\
        OmniPhysGS~\cite{lin2025omniphysgs} &  \underline{0.622}  &  0.723  & 0.402 & 2.19 & 2.55 & 1.92 \\
        PhysGen~\cite{liu2024physgen} &  0.547  &  0.808  & 0.550 & 2.17 & 2.07 & 2.76  \\
        PhysGen3D~\cite{chen2025physgen3d} &  0.621  &   \underline{0.879} & 0.589  & 2.07 & 2.27 & 2.06\\
        \midrule
        \textbf{Ours}  & \textbf{0.627}   & \textbf{0.885}   &  \textbf{0.636} & \textbf{3.47}	& \textbf{3.52} &	\textbf{3.53}
        \\
        \bottomrule
    \end{tabularx}%
    \vspace{-0.4cm}
    \label{tab:subjective}
\end{table}

\begin{figure*}[t]
    \centering
    \includegraphics[width=1\linewidth]{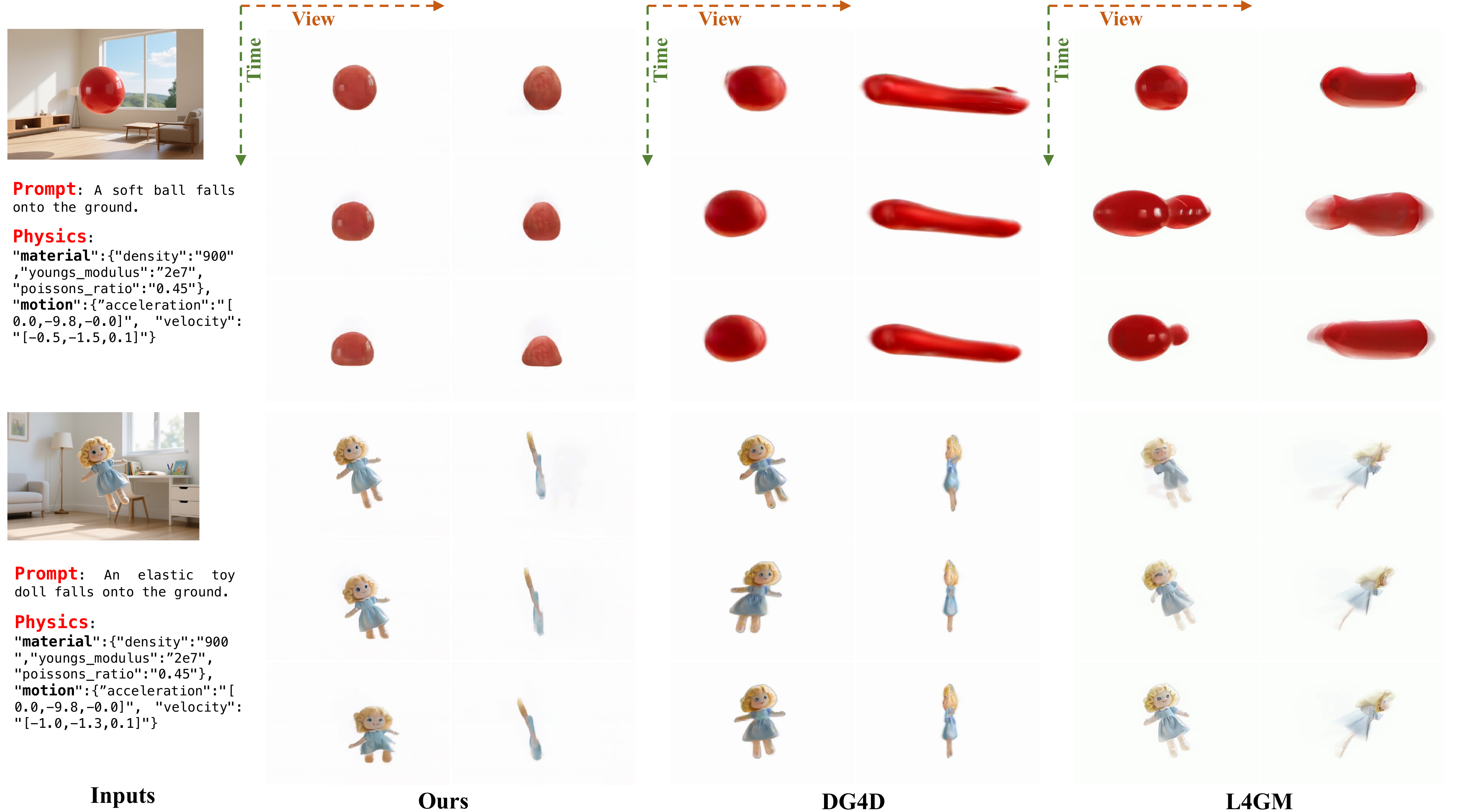}
    \vspace{-0.5cm}
     \caption{\textbf{Qualitative comparison of orthogonal-view video synthesis results}.}
    \label{fig:com_4view}
    \vspace{-0.7cm}
\end{figure*}

\paragraph{Orthogonal-view video quality comparison.}
To evaluate the quality of the generated orthogonal-view videos, we compare our method with existing 4D generation approaches, including DG4D~\citep{ren2023dreamgaussian4d}, $\text{Diffusion}^2$~\cite{yang2024diffusion}, and L4GM~\citep{ren2024l4gm}. For quantitative evaluation, we adopt VBench metrics to assess video quality from different views. As shown in Table~\ref{tab:4view}, our method outperforms the competing approaches in terms of both motion consistency and visual quality. For qualitative comparison, representative results are shown in Fig.~\ref{fig:com_4view}. Our method generates motion that follows physical principles, whereas existing 4D generation methods often fail to do so. Moreover, our approach effectively captures orthogonal-view structure, producing consistent results from novel viewpoints.

\begin{table}[t]
    \centering
    \caption{\textbf{Evaluation of foreground orthogonal-view video synthesis.}}
    \vspace{-0.1cm}
    \setlength{\tabcolsep}{10pt}
    \renewcommand{\arraystretch}{1.0} 
    \begin{tabularx}{\linewidth}{@{} lccccccc @{}}
    \toprule
    \multirow{2}{*}{Method} & \multicolumn{3}{c}{View1} & \multicolumn{3}{c}{View2} \\
    \cmidrule(lr){2-4} \cmidrule(lr){5-7}
    & Motion↑ & Subject↑  & Image↑  & Motion↑ & Subject↑  & Image↑   \\ \midrule
        DG4D~\cite{ren2023dreamgaussian4d} & \underline{0.986}  &  0.892 & 0.268   &  0.987 & 0.883 & 0.260  \\
        $\text{Diffusion}^2$~\cite{yang2024diffusion} & 0.970 & 0.637 & \underline{0.371} & 0.979 & 0.666 & \underline{0.362}  \\
        L4GM~\cite{ren2024l4gm}  &  0.994 &  \underline{0.898} &  0.368 & \underline{0.992} & \underline{0.899} & 0.305  \\
        \midrule
        \textbf{Ours}  & \textbf{0.995}  & \textbf{0.912}  &  \textbf{0.389}  & \textbf{0.994} & \textbf{0.901} & \textbf{0.368}  \\
        \bottomrule
    \end{tabularx}%
    \label{tab:4view}
    \vspace{-0.4cm}
\end{table}

\subsection{Ablation Study}
\begin{figure}[t]
    \centering
    \begin{minipage}[t]{0.45\linewidth}
        \vspace{0pt}
        \centering
        \captionof{table}{\textbf{Ablation study of attention modules in the orthogonal-view generation stage}.}
        \setlength{\tabcolsep}{3pt}
        \renewcommand{\arraystretch}{1.0}
        \small
        \begin{tabularx}{\linewidth}{@{} lccc @{}}
        \toprule
            Method & Motion↑  & Image↑ & Physical↑ \\ \midrule
            Baseline* & 0.994 & 0.594 & 0.636\\
            + Physics-Attn & 0.994 & 0.613 & 0.682 \\
            + Cross-View Attn & 0.995 & 0.648 & 0.689 \\
            + Temporal-Attn & \textbf{0.996} & \textbf{0.655} & \textbf{0.691} \\
            \bottomrule
        \end{tabularx}
        \label{tab:ablation_4view}
    \end{minipage}
    \hfill
    \begin{minipage}[t]{0.49\linewidth}
        \vspace{0pt}
        \centering
        \includegraphics[width=\linewidth]{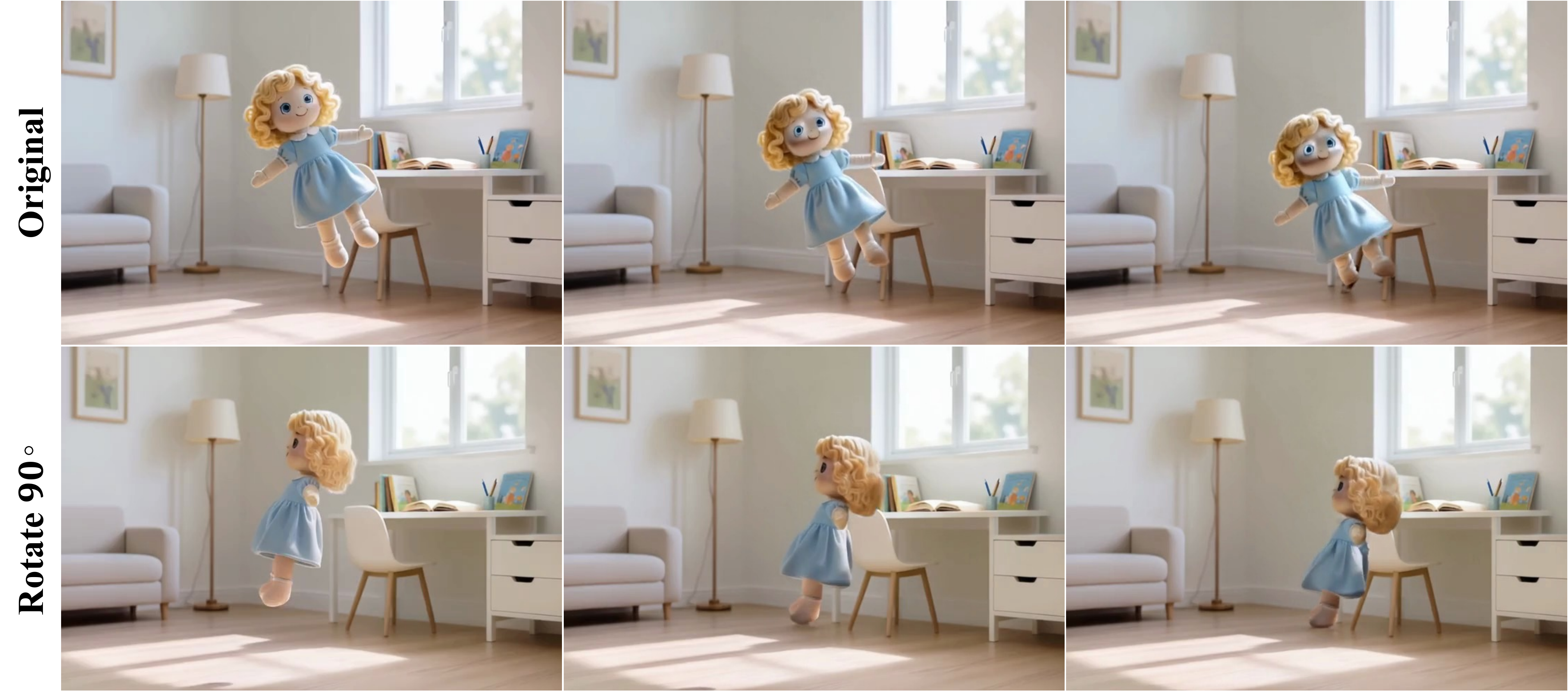}
        \captionof{figure}{\textbf{Results of object editing}.}
        \label{fig:object-4view}
    \end{minipage}
    \vspace{-0.4cm}
\end{figure}

We conduct ablation studies on the orthogonal-view generation stage of OrthoPhys to verify the effectiveness of its key components. Specifically, we incrementally incorporate physics-aware attention (Phys-Attn), geometry-enhanced cross-view attention (Cross-view Attn), and temporal attention (Temporal-Attn). We evaluate results using two VBench metrics, namely Motion and Image, together with a physical realism score by GPT-4o. As shown in Table~\ref{tab:ablation_4view}, introducing physics-aware attention leads to a significant improvement in physical realism. Adding geometry-enhanced cross-view attention further enhances motion smoothness and image quality, while temporal attention provides additional gains in motion consistency. The qualitative ablation comparison is shown in Appendix Fig.~\ref{fig:ablation_4view}. Additional ablation study on condition videos is detailed in Appendix Fig.~\ref{fig:ablation_con}.

\section{Downstream Applications}

While our primary focus is video generation rather than full 4D modeling, the orthogonal four-view foreground videos produced by our method naturally enable several downstream applications.
As a demonstration, we show that the generated four-view foreground videos are well-suited for video editing. As illustrated in Fig.~\ref{fig:object-4view}, we modify the rotation of the foreground object and leverage the foreground video segment from one of the generated views to guide the synthesis of an edited video.
Additionally, although not a central goal of our method, for scenes with simple backgrounds, our approach can generate four-view videos with background, producing spatially consistent foreground–background renderings across views, as shown in Appendix Fig.~\ref{fig:result_4view_bg}.

\section{Conclusion}
In this work, we propose OrthoPhys, a two-stage framework for physically plausible video generation with orthogonal-view geometry guidance. OrthoPhys introduces physics-aware and geometry-enhanced inductive biases to improve motion consistency and physical plausibility without relying on explicit 3D representation. The framework combines physics-conditioned orthogonal-view motion modeling with background-aware video synthesis, enabling controllable video generation. Extensive experiments demonstrate that OrthoPhys achieves superior motion coherence, visual quality, and physical realism compared to existing methods. In addition, we introduce a large-scale physics-oriented multi-view video dataset to facilitate future research.

{\small
\bibliography{main}

@inproceedings{yangcogvideox,
  title={CogVideoX: Text-to-Video Diffusion Models with An Expert Transformer},
  author={Yang, Zhuoyi and Teng, Jiayan and Zheng, Wendi and Ding, Ming and Huang, Shiyu and Xu, Jiazheng and Yang, Yuanming and Hong, Wenyi and Zhang, Xiaohan and Feng, Guanyu and others},
year={2025},
  booktitle={The Thirteenth International Conference on Learning Representations}
}

@article{wan2025wan,
  title={Wan: Open and advanced large-scale video generative models},
  author={Wan, Team and Wang, Ang and Ai, Baole and Wen, Bin and Mao, Chaojie and Xie, Chen-Wei and Chen, Di and Yu, Feiwu and Zhao, Haiming and Yang, Jianxiao and others},
  journal={arXiv preprint arXiv:2503.20314},
  year={2025}
}

@article{gillman2025force,
  title={Force Prompting: Video Generation Models Can Learn and Generalize Physics-based Control Signals},
  author={Gillman, Nate and Herrmann, Charles and Freeman, Michael and Aggarwal, Daksh and Luo, Evan and Sun, Deqing and Sun, Chen},
  journal={arXiv preprint arXiv:2505.19386},
  year={2025}
}

@article{lin2025omniphysgs,
  title={OmniphysGS: 3d constitutive gaussians for general physics-based dynamics generation},
  author={Lin, Yuchen and Lin, Chenguo and Xu, Jianjin and Mu, Yadong},
  journal={arXiv preprint arXiv:2501.18982},
  year={2025}
}

@inproceedings{liu2024physgen,
  title={Physgen: Rigid-body physics-grounded image-to-video generation},
  author={Liu, Shaowei and Ren, Zhongzheng and Gupta, Saurabh and Wang, Shenlong},
  booktitle={European Conference on Computer Vision},
  pages={360--378},
  year={2024}
}

@inproceedings{chen2025physgen3d,
  title={Physgen3d: Crafting a miniature interactive world from a single image},
  author={Chen, Boyuan and Jiang, Hanxiao and Liu, Shaowei and Gupta, Saurabh and Li, Yunzhu and Zhao, Hao and Wang, Shenlong},
  booktitle={Proceedings of the Computer Vision and Pattern Recognition Conference},
  pages={6178--6189},
  year={2025}
}

@article{ren2023dreamgaussian4d,
  title={Dreamgaussian4d: Generative 4d gaussian splatting},
  author={Ren, Jiawei and Pan, Liang and Tang, Jiaxiang and Zhang, Chi and Cao, Ang and Zeng, Gang and Liu, Ziwei},
  journal={arXiv preprint arXiv:2312.17142},
  year={2023}
}

@article{ren2024l4gm,
  title={L4gm: Large 4d gaussian reconstruction model},
  author={Ren, Jiawei and Xie, Cheng and Mirzaei, Ashkan and Kreis, Karsten and Liu, Ziwei and Torralba, Antonio and Fidler, Sanja and Kim, Seung Wook and Ling, Huan and others},
  journal={Advances in Neural Information Processing Systems},
  volume={37},
  pages={56828--56858},
  year={2024}
}

@article{wu2025qwen,
  title={Qwen-image technical report},
  author={Wu, Chenfei and Li, Jiahao and Zhou, Jingren and Lin, Junyang and Gao, Kaiyuan and Yan, Kun and Yin, Sheng-ming and Bai, Shuai and Xu, Xiao and Chen, Yilei and others},
  journal={arXiv preprint arXiv:2508.02324},
  year={2025}
}

@article{ren2024grounded,
  title={Grounded sam: Assembling open-world models for diverse visual tasks},
  author={Ren, Tianhe and Liu, Shilong and Zeng, Ailing and Lin, Jing and Li, Kunchang and Cao, He and Chen, Jiayu and Huang, Xinyu and Chen, Yukang and Yan, Feng and others},
  journal={arXiv preprint arXiv:2401.14159},
  year={2024}
}

@inproceedings{xiang2025structured,
  title={Structured 3d latents for scalable and versatile 3d generation},
  author={Xiang, Jianfeng and Lv, Zelong and Xu, Sicheng and Deng, Yu and Wang, Ruicheng and Zhang, Bowen and Chen, Dong and Tong, Xin and Yang, Jiaolong},
  booktitle={Proceedings of the Computer Vision and Pattern Recognition Conference},
  pages={21469--21480},
  year={2025}
}

@inproceedings{huang2024vbench,
  title={Vbench: Comprehensive benchmark suite for video generative models},
  author={Huang, Ziqi and He, Yinan and Yu, Jiashuo and Zhang, Fan and Si, Chenyang and Jiang, Yuming and Zhang, Yuanhan and Wu, Tianxing and Jin, Qingyang and Chanpaisit, Nattapol and others},
  booktitle={Proceedings of the IEEE/CVF Conference on Computer Vision and Pattern Recognition},
  pages={21807--21818},
  year={2024}
}

@article{duan2025worldscore,
  title={Worldscore: A unified evaluation benchmark for world generation},
  author={Duan, Haoyi and Yu, Hong-Xing and Chen, Sirui and Fei-Fei, Li and Wu, Jiajun},
  journal={arXiv preprint arXiv:2504.00983},
  year={2025}
}

@article{ho2022video,
  title={Video diffusion models},
  author={Ho, Jonathan and Salimans, Tim and Gritsenko, Alexey and Chan, William and Norouzi, Mohammad and Fleet, David J},
  journal={Advances in neural information processing systems},
  volume={35},
  pages={8633--8646},
  year={2022}
}

@article{blattmann2023stable,
  title={Stable video diffusion: Scaling latent video diffusion models to large datasets},
  author={Blattmann, Andreas and Dockhorn, Tim and Kulal, Sumith and Mendelevitch, Daniel and Kilian, Maciej and Lorenz, Dominik and Levi, Yam and English, Zion and Voleti, Vikram and Letts, Adam and others},
  journal={arXiv preprint arXiv:2311.15127},
  year={2023}
}

@article{kong2024hunyuanvideo,
  title={Hunyuanvideo: A systematic framework for large video generative models},
  author={Kong, Weijie and Tian, Qi and Zhang, Zijian and Min, Rox and Dai, Zuozhuo and Zhou, Jin and Xiong, Jiangfeng and Li, Xin and Wu, Bo and Zhang, Jianwei and others},
  journal={arXiv preprint arXiv:2412.03603},
  year={2024}
}

@article{fu20243dtrajmaster,
  title={3dtrajmaster: Mastering 3d trajectory for multi-entity motion in video generation},
  author={Fu, Xiao and Liu, Xian and Wang, Xintao and Peng, Sida and Xia, Menghan and Shi, Xiaoyu and Yuan, Ziyang and Wan, Pengfei and Zhang, Di and Lin, Dahua},
  journal={arXiv preprint arXiv:2412.07759},
  year={2024}
}

@article{he2024cameractrl,
  title={Cameractrl: Enabling camera control for text-to-video generation},
  author={He, Hao and Xu, Yinghao and Guo, Yuwei and Wetzstein, Gordon and Dai, Bo and Li, Hongsheng and Yang, Ceyuan},
  journal={arXiv preprint arXiv:2404.02101},
  year={2024}
}

@inproceedings{geng2025motion,
  title={Motion prompting: Controlling video generation with motion trajectories},
  author={Geng, Daniel and Herrmann, Charles and Hur, Junhwa and Cole, Forrester and Zhang, Serena and Pfaff, Tobias and Lopez-Guevara, Tatiana and Aytar, Yusuf and Rubinstein, Michael and Sun, Chen and others},
  booktitle={Proceedings of the Computer Vision and Pattern Recognition Conference},
  pages={1--12},
  year={2025}
}

@inproceedings{gu2025diffusion,
  title={Diffusion as shader: 3d-aware video diffusion for versatile video generation control},
  author={Gu, Zekai and Yan, Rui and Lu, Jiahao and Li, Peng and Dou, Zhiyang and Si, Chenyang and Dong, Zhen and Liu, Qifeng and Lin, Cheng and Liu, Ziwei and others},
  booktitle={Proceedings of the Special Interest Group on Computer Graphics and Interactive Techniques Conference Conference Papers},
  pages={1--12},
  year={2025}
}

@inproceedings{burgert2025go,
  title={Go-with-the-flow: Motion-controllable video diffusion models using real-time warped noise},
  author={Burgert, Ryan and Xu, Yuancheng and Xian, Wenqi and Pilarski, Oliver and Clausen, Pascal and He, Mingming and Ma, Li and Deng, Yitong and Li, Lingxiao and Mousavi, Mohsen and others},
  booktitle={Proceedings of the Computer Vision and Pattern Recognition Conference},
  pages={13--23},
  year={2025}
}

@article{chen2025stance,
  title={STANCE: Motion Coherent Video Generation Via Sparse-to-Dense Anchored Encoding},
  author={Chen, Zhifei and Xu, Tianshuo and Wu, Leyi and Wang, Luozhou and Yan, Dongyu and You, Zihan and Luo, Wenting and Zhang, Guo and Chen, Yingcong},
  journal={arXiv preprint arXiv:2510.14588},
  year={2025}
}

@article{romero2025learning,
  title={Learning to Generate Object Interactions with Physics-Guided Video Diffusion},
  author={Romero, David and Bermudez, Ariana and Li, Hao and Pizzati, Fabio and Laptev, Ivan},
  journal={arXiv preprint arXiv:2510.02284},
  year={2025}
}

@inproceedings{li2024generative,
  title={Generative image dynamics},
  author={Li, Zhengqi and Tucker, Richard and Snavely, Noah and Holynski, Aleksander},
  booktitle={Proceedings of the IEEE/CVF Conference on Computer Vision and Pattern Recognition},
  pages={24142--24153},
  year={2024}
}

@inproceedings{liang2024flowvid,
  title={Flowvid: Taming imperfect optical flows for consistent video-to-video synthesis},
  author={Liang, Feng and Wu, Bichen and Wang, Jialiang and Yu, Licheng and Li, Kunpeng and Zhao, Yinan and Misra, Ishan and Huang, Jia-Bin and Zhang, Peizhao and Vajda, Peter and others},
  booktitle={Proceedings of the IEEE/CVF Conference on Computer Vision and Pattern Recognition},
  pages={8207--8216},
  year={2024}
}

@inproceedings{jin2025flovd,
  title={Flovd: Optical flow meets video diffusion model for enhanced camera-controlled video synthesis},
  author={Jin, Wonjoon and Dai, Qi and Luo, Chong and Baek, Seung-Hwan and Cho, Sunghyun},
  booktitle={Proceedings of the Computer Vision and Pattern Recognition Conference},
  pages={2040--2049},
  year={2025}
}

@inproceedings{lv2024gpt4motion,
  title={Gpt4motion: Scripting physical motions in text-to-video generation via blender-oriented gpt planning},
  author={Lv, Jiaxi and Huang, Yi and Yan, Mingfu and Huang, Jiancheng and Liu, Jianzhuang and Liu, Yifan and Wen, Yafei and Chen, Xiaoxin and Chen, Shifeng},
  booktitle={Proceedings of the IEEE/CVF conference on computer vision and pattern recognition},
  pages={1430--1440},
  year={2024}
}

@inproceedings{xue2025phyt2v,
  title={Phyt2v: Llm-guided iterative self-refinement for physics-grounded text-to-video generation},
  author={Xue, Qiyao and Yin, Xiangyu and Yang, Boyuan and Gao, Wei},
  booktitle={Proceedings of the Computer Vision and Pattern Recognition Conference},
  pages={18826--18836},
  year={2025}
}

@article{wang2025wisa,
  title={Wisa: World simulator assistant for physics-aware text-to-video generation},
  author={Wang, Jing and Ma, Ao and Cao, Ke and Zheng, Jun and Zhang, Zhanjie and Feng, Jiasong and Liu, Shanyuan and Ma, Yuhang and Cheng, Bo and Leng, Dawei and others},
  journal={arXiv preprint arXiv:2503.08153},
  year={2025}
}

@article{bansal2024videophy,
  title={Videophy: Evaluating physical commonsense for video generation},
  author={Bansal, Hritik and Lin, Zongyu and Xie, Tianyi and Zong, Zeshun and Yarom, Michal and Bitton, Yonatan and Jiang, Chenfanfu and Sun, Yizhou and Chang, Kai-Wei and Grover, Aditya},
  journal={arXiv preprint arXiv:2406.03520},
  year={2024}
}

@article{meng2024towards,
  title={Towards world simulator: Crafting physical commonsense-based benchmark for video generation},
  author={Meng, Fanqing and Liao, Jiaqi and Tan, Xinyu and Shao, Wenqi and Lu, Quanfeng and Zhang, Kaipeng and Cheng, Yu and Li, Dianqi and Qiao, Yu and Luo, Ping},
  journal={arXiv preprint arXiv:2410.05363},
  year={2024}
}

@article{kerbl20233d,
  title={3D Gaussian splatting for real-time radiance field rendering.},
  author={Kerbl, Bernhard and Kopanas, Georgios and Leimk{\"u}hler, Thomas and Drettakis, George},
  journal={ACM Trans. Graph.},
  volume={42},
  number={4},
  pages={139--1},
  year={2023}
}

@article{feng2024gaussian,
  title={Gaussian splashing: Dynamic fluid synthesis with gaussian splatting},
  author={Feng, Yutao and Feng, Xiang and Shang, Yintong and Jiang, Ying and Yu, Chang and Zong, Zeshun and Shao, Tianjia and Wu, Hongzhi and Zhou, Kun and Jiang, Chenfanfu and others},
  journal={CoRR},
  year={2024}
}

@inproceedings{gao2025fluidnexus,
  title={FluidNexus: 3D fluid reconstruction and prediction from a single video},
  author={Gao, Yue and Yu, Hong-Xing and Zhu, Bo and Wu, Jiajun},
  booktitle={Proceedings of the Computer Vision and Pattern Recognition Conference},
  pages={26091--26101},
  year={2025}
}

@inproceedings{zhao2025physsplat,
  title={PhysSplat: Efficient Physics Simulation for 3D Scenes via MLLM-Guided Gaussian Splatting},
  author={Zhao, Haoyu and Wang, Hao and Zhao, Xingyue and Fei, Hao and Wang, Hongqiu and Long, Chengjiang and Zou, Hua},
  booktitle={Proceedings of the IEEE/CVF International Conference on Computer Vision},
  pages={5242--5252},
  year={2025}
}

@inproceedings{mao2025live,
  title={LIVE-GS: LLM Powers Interactive VR by Enhancing Gaussian Splatting},
  author={Mao, Haotian and Xu, Zhuoxiong and Wei, Siyue and Quan, Yule and Deng, Nianchen and Yang, Xubo},
  booktitle={2025 IEEE Conference on Virtual Reality and 3D User Interfaces Abstracts and Workshops (VRW)},
  pages={1234--1235},
  year={2025},
  organization={IEEE}
}

@inproceedings{huang2025dreamphysics,
  title={DreamPhysics: Learning Physics-Based 3D Dynamics with Video Diffusion Priors},
  author={Huang, Tianyu and Zhang, Haoze and Zeng, Yihan and Zhang, Zhilu and Li, Hui and Zuo, Wangmeng and Lau, Rynson WH},
  booktitle={Proceedings of the AAAI Conference on Artificial Intelligence},
  pages={3733--3741},
  year={2025}
}

@article{hu2019taichi,
  title={Taichi: a language for high-performance computation on spatially sparse data structures},
  author={Hu, Yuanming and Li, Tzu-Mao and Anderson, Luke and Ragan-Kelley, Jonathan and Durand, Fr{\'e}do},
  journal={ACM Transactions on Graphics (TOG)},
  pages={1--16},
  year={2019},
  publisher={ACM New York, NY, USA}
}

@inproceedings{xie2024physgaussian,
  title={Physgaussian: Physics-integrated 3d gaussians for generative dynamics},
  author={Xie, Tianyi and Zong, Zeshun and Qiu, Yuxing and Li, Xuan and Feng, Yutao and Yang, Yin and Jiang, Chenfanfu},
  booktitle={Proceedings of the IEEE/CVF Conference on Computer Vision and Pattern Recognition},
  pages={4389--4398},
  year={2024}
}

@inproceedings{zhang2024physdreamer,
  title={Physdreamer: Physics-based interaction with 3d objects via video generation},
  author={Zhang, Tianyuan and Yu, Hong-Xing and Wu, Rundi and Feng, Brandon Y and Zheng, Changxi and Snavely, Noah and Wu, Jiajun and Freeman, William T},
  booktitle={European Conference on Computer Vision},
  pages={388--406},
  year={2024}
}

@article{poole2022dreamfusion,
  title={Dreamfusion: Text-to-3d using 2d diffusion},
  author={Poole, Ben and Jain, Ajay and Barron, Jonathan T and Mildenhall, Ben},
  journal={arXiv preprint arXiv:2209.14988},
  year={2022}
}

@inproceedings{dong2024memflow,
  title={Memflow: Optical flow estimation and prediction with memory},
  author={Dong, Qiaole and Fu, Yanwei},
  booktitle={Proceedings of the IEEE/CVF Conference on Computer Vision and Pattern Recognition},
  pages={19068--19078},
  year={2024}
}

@article{yang2024depth,
  title={Depth anything v2},
  author={Yang, Lihe and Kang, Bingyi and Huang, Zilong and Zhao, Zhen and Xu, Xiaogang and Feng, Jiashi and Zhao, Hengshuang},
  journal={Advances in Neural Information Processing Systems},
  volume={37},
  pages={21875--21911},
  year={2024}
}

@article{zheng2024open,
  title={Open-sora: Democratizing efficient video production for all},
  author={Zheng, Zangwei and Peng, Xiangyu and Yang, Tianji and Shen, Chenhui and Li, Shenggui and Liu, Hongxin and Zhou, Yukun and Li, Tianyi and You, Yang},
  journal={arXiv preprint arXiv:2412.20404},
  year={2024}
}

@inproceedings{lin2025phys4dgen,
  title={Phys4DGen: Physics-Compliant 4D Generation with Multi-Material Composition Perception},
  author={Lin, Jiajing and Wang, Zhenzhong and Xu, Dejun and Jiang, Shu and Gong, Yunpeng and Jiang, Min},
  booktitle={Proceedings of the 33rd ACM International Conference on Multimedia},
  pages={10398--10407},
  year={2025}
}

@article{liu2024physics3d,
  title={Physics3d: Learning physical properties of 3d gaussians via video diffusion},
  author={Liu, Fangfu and Wang, Hanyang and Yao, Shunyu and Zhang, Shengjun and Zhou, Jie and Duan, Yueqi},
  journal={arXiv preprint arXiv:2406.04338},
  year={2024}
}

@article{nan2024openvid,
  title={Openvid-1m: A large-scale high-quality dataset for text-to-video generation},
  author={Nan, Kepan and Xie, Rui and Zhou, Penghao and Fan, Tiehan and Yang, Zhenheng and Chen, Zhijie and Li, Xiang and Yang, Jian and Tai, Ying},
  journal={arXiv preprint arXiv:2407.02371},
  year={2024}
}

@inproceedings{yang2024diffusion,
  title={Diffusion²: Dynamic 3D Content Generation via Score Composition of Video and Multi-view Diffusion Models},
  author={Yang, Zeyu and Pan, Zijie and Gu, Chun and Zhang, Li},
  booktitle={International Conference on Learning Representations (ICLR)},
  year={2025}
}

@inproceedings{wu2024perception,
  title={Perception-oriented video frame interpolation via asymmetric blending},
  author={Wu, Guangyang and Tao, Xin and Li, Changlin and Wang, Wenyi and Liu, Xiaohong and Zheng, Qingqing},
  booktitle={Proceedings of the IEEE/CVF Conference on Computer Vision and Pattern Recognition},
  pages={2753--2762},
  year={2024}
}
}
\newpage
\appendix
\onecolumn



\section{Additional Experiments}\label{additional_exp}

\subsection{Baseline Details}\label{sec:app_baseline_details}
We compare OrthoPhys with the baselines reported in Table~\ref{tab:com} and Table~\ref{tab:subjective}. The comparison is organized into two categories: physics-engine-based methods and learning-based video generation models. All methods are evaluated on the same benchmark samples and with the same metric pipelines described in Sec.~\ref{sec:exp_setup}; no baseline is given access to the orthogonal-view geometry guidance or foreground motion priors produced by OrthoPhys.

\paragraph{Physics-engine-based baselines.}
\begin{itemize}[leftmargin=*,topsep=2pt,itemsep=2pt]
    \item \textbf{PhysGen}~\citep{liu2024physgen}: a physics-grounded generation method that uses physical simulation to model object dynamics. We include it as a representative baseline for simulation-based physically plausible video generation.
    \item \textbf{PhysGen3D}~\citep{chen2025physgen3d}: a 3D-aware physics-based video generation baseline that combines physical simulation with explicit 3D representations. It serves as a strong reference for methods that improve physical plausibility through scene-level physical modeling.
    \item \textbf{OmniPhysGS}~\citep{lin2025omniphysgs}: a Gaussian-splatting-based physical simulation method for dynamic scene generation. We include it to compare against approaches that explicitly simulate object dynamics in a 3D scene representation.
\end{itemize}

\paragraph{Learning-based video generation baselines.}
\begin{itemize}[leftmargin=*,topsep=2pt,itemsep=2pt]
    \item \textbf{CogVideoX}~\citep{yangcogvideox}: a strong open video generation backbone with an expert-transformer design. It serves as a general-purpose video generation baseline.
    \item \textbf{Wan}~\citep{wan2025wan}: a recent high-quality video generation model used to assess whether strong visual generation capacity alone can produce physically plausible motion and object-background interactions.
    \item \textbf{Force Prompting}~\citep{gillman2025force}: a physics-aware prompting method that augments video generation with explicit physical priors. We include it to compare against learning-based generation enhanced through textual or prompt-level physical guidance.
\end{itemize}

\subsection{Additional Quantitative Physical Evaluations.}\label{sec:app_additional_physical_eval}

\paragraph{VideoPhy-2 evaluation.}
We further evaluate all comparison methods on VideoPhy-2~\citep{bansal2024videophy}, a complementary benchmark for physical commonsense in video generation. As shown in Table~\ref{tab:app_videophy2}, OrthoPhys achieves the best physical plausibility and semantic consistency scores among all evaluated methods, indicating that the gains observed on our benchmark also transfer to an external physical evaluation protocol.

\begin{table}[H]
    \centering
    \caption{\textbf{Quantitative comparisons on VideoPhy-2}.}
    \setlength{\tabcolsep}{18pt}
    \renewcommand{\arraystretch}{1.05}
    \begin{tabular}{@{} lcc @{}}
    \toprule
        Method & Physical↑ & Semantic↑ \\ \midrule
        CogVideoX-5B & \underline{3.06} & 3.33 \\
        Wan2.2-5B & 2.91 & 3.45 \\
        ForcePrompt & 2.72 & 3.27 \\
        OmniPhysGS & 2.21 & 3.53 \\
        PhysGen & 2.69 & \underline{3.63} \\
        PhysGen3D & 2.73 & 3.57 \\
        \textbf{Ours} & \textbf{3.12} & \textbf{3.83} \\
        \bottomrule
    \end{tabular}
    \label{tab:app_videophy2}
\end{table}

\paragraph{Controlled physical controllability.}
To provide a more direct quantitative validation of physical controllability, we construct a controlled synthetic paired evaluation set generated by the physics simulator. Since the available supervision consists of predicted videos paired with reference motion videos, we report three proxy physical metrics that are closely tied to physical behavior: trajectory RMSE, contact timing error, and deformation-trend error. These metrics quantify motion alignment, contact-event consistency, and deformation consistency, respectively. Contact timing error measures the deviation in the first-contact frame, while deformation-trend error measures the discrepancy between predicted and reference deformation evolution over time.

On our constructed 100-sample evaluation set, OrthoPhys consistently outperforms DG4D across all three physical motion metrics, as shown in Table~\ref{tab:app_controlled_physical}. Although these are proxy metrics rather than full physical-state errors, they provide a direct quantitative link between physical inputs and generated motion.

\begin{table}[H]
    \centering
    \caption{\textbf{Quantitative evaluation on the controlled synthetic evaluation set}. PSNR is reported as a reference reconstruction metric, while Traj. RMSE, Contact Timing Err., and Def. Trend Err. are proxy physical metrics.}
    \setlength{\tabcolsep}{9pt}
    \renewcommand{\arraystretch}{1.05}
    \begin{tabular}{@{} lcccc @{}}
    \toprule
        Method & PSNR↑ & Traj. RMSE↓ & Contact Timing Err.↓ & Def. Trend Err.↓ \\ \midrule
        DG4D & 18.32 & 0.1203 & 19.59 & 36.51 \\
        \textbf{Ours} & \textbf{22.24} & \textbf{0.0148} & \textbf{18.03} & \textbf{11.11} \\
        \bottomrule
    \end{tabular}
    \label{tab:app_controlled_physical}
\end{table}

\begin{figure}[t]
    \centering
    \includegraphics[width=1\linewidth]{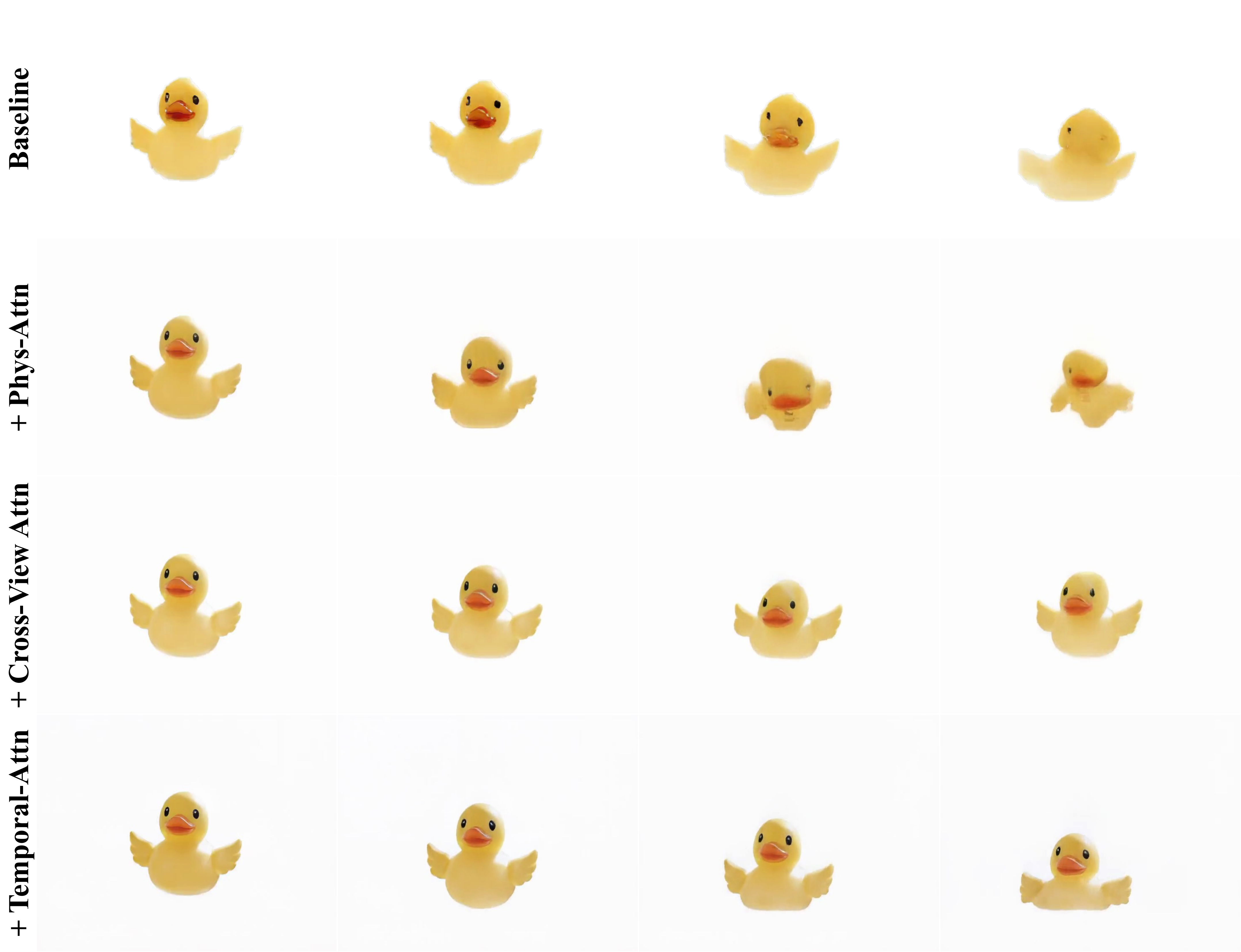}
    \caption{\textbf{Qualitative ablation results}. We compare the baseline with variants that incorporate individual components, including Phys-Attn, Cross-View Attn, and Temporal-Attn. Each row corresponds to one setting.}
    \label{fig:ablation_4view}
\end{figure}

\subsection{Ablation Study of Attention Modules.}
In the main manuscript, we report quantitative ablation results in Table~\ref{tab:ablation_4view} to evaluate the contribution of different attention modules. Here, we further provide qualitative ablation results in Fig.~\ref{fig:ablation_4view} to offer visual insights into their effects on orthogonal foreground video generation. As shown in Fig.~\ref{fig:ablation_4view}, progressively introducing physics-aware attention, geometry-enhanced cross-view attention, and temporal attention leads to more coherent motion patterns and improved visual quality. These qualitative results complement the quantitative analysis and further validate the effectiveness of each proposed component.

\subsection{Ablation Study of Condition Videos.}
\begin{figure}[t]
    \centering
    \includegraphics[width=1\linewidth]{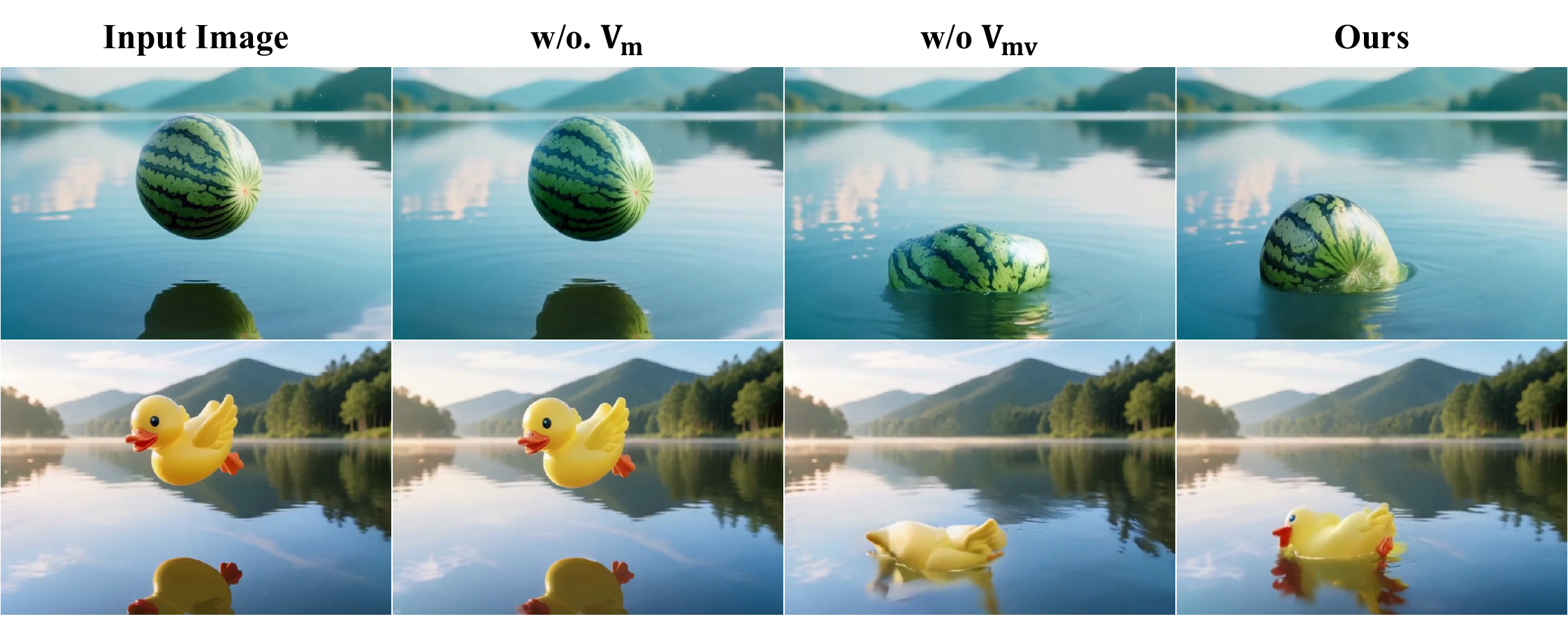}
    \caption{\textbf{Ablation study on conditional videos}. The left column shows the input single image, and the three right columns each display one representative frame from the generated RGB video.}
    \label{fig:ablation_con}
\end{figure}

We employ two types of video conditions to guide the orthogonal-view generation process. To evaluate the effectiveness of these conditions, we conduct an ablation study with qualitative comparisons shown in Fig.~\ref{fig:ablation_con}. The foreground mask video condition provides explicit guidance for foreground object motion, while the orthogonal-view video condition implicitly encodes 3D structural information, leading to more spatially consistent four-view video generation.

\subsection{Supplementary Ablation Studies.}
We provide additional ablation studies to isolate finer-grained design choices in OrthoPhys. Unless otherwise specified, we report Motion and Image scores from VBench and the Physical score from GPT-4o evaluation, following the metrics used in the main ablation study.

\paragraph{Finer-grained pipeline components.}
Table~\ref{tab:app_ablation_components} evaluates several intermediate design choices beyond the major attention modules, including the mask prior, orthogonal-view prior, gated fusion, and depth-confidence weighting. Removing any of these components degrades the final performance, with a particularly clear drop in physical plausibility. This indicates that the gains come from a compact set of necessary design choices rather than arbitrary engineering details. Frame interpolation, optical-flow extraction, and flow-conditioned synthesis are supporting implementation steps: they align the guidance sequence, extract foreground motion, and transfer motion cues from the first-stage orthogonal-view results to full-scene synthesis without directly propagating appearance artifacts.

\begin{table}[H]
    \centering
    \caption{\textbf{Ablation on finer-grained design choices in OrthoPhys}.}
    \setlength{\tabcolsep}{8pt}
    \renewcommand{\arraystretch}{1.0}
    \begin{tabular}{@{} lccc @{}}
    \toprule
        Method & Motion↑ & Image↑ & Physical↑ \\ \midrule
        Ours (w/o mask prior) & 0.991 & 0.646 & 0.609 \\
        Ours (w/o orthogonal-view prior) & 0.992 & 0.637 & 0.625 \\
        Ours (w/o gated fusion) & 0.995 & 0.651 & 0.648 \\
        Ours (w/o depth-confidence weighting) & 0.994 & 0.642 & 0.643 \\
        \textbf{Ours} & \textbf{0.996} & \textbf{0.655} & \textbf{0.691} \\
        \bottomrule
    \end{tabular}
    \label{tab:app_ablation_components}
\end{table}

\paragraph{Two-stage design.}
We compare the full OrthoPhys pipeline with a single-stage variant that uses the same physics conditioning but does not generate orthogonal-view foreground videos. As shown in Table~\ref{tab:app_ablation_twostage}, removing the orthogonal-view foreground stage consistently reduces all metrics. This supports the importance of decomposing physically grounded foreground motion generation from the final background-aware synthesis stage.

\begin{table}[H]
    \centering
    \caption{\textbf{Ablation study on the two-stage design}.}
    \setlength{\tabcolsep}{8pt}
    \renewcommand{\arraystretch}{1.0}
    \begin{tabular}{@{} lccc @{}}
    \toprule
        Method & Motion↑ & Image↑ & Physical↑ \\ \midrule
        Ours (w/o orthogonal-view constraints) & 0.983 & 0.636 & 0.628 \\
        \textbf{Ours} & \textbf{0.996} & \textbf{0.655} & \textbf{0.691} \\
        \bottomrule
    \end{tabular}
    \label{tab:app_ablation_twostage}
\end{table}

\paragraph{Geometry-enhanced cross-view attention.}
Table~\ref{tab:app_ablation_geometry_crossview} further analyzes the geometry-enhanced cross-view attention module. Cross-view attention alone already improves over removing cross-view interaction, while the full geometry-enhanced variant brings additional gains by better capturing spatial dependencies across synchronized orthogonal views.

\begin{table}[H]
    \centering
    \caption{\textbf{Ablation study on geometry-enhanced cross-view attention}.}
    \setlength{\tabcolsep}{8pt}
    \renewcommand{\arraystretch}{1.0}
    \begin{tabular}{@{} lccc @{}}
    \toprule
        Method & Motion↑ & Image↑ & Physical↑ \\ \midrule
        Ours (w/o cross-view) & 0.994 & 0.594 & 0.636 \\
        Ours (only Cross-Attn) & 0.994 & 0.640 & 0.678 \\
        \textbf{Ours (w/ Geometry-enhanced Attn)} & \textbf{0.996} & \textbf{0.655} & \textbf{0.691} \\
        \bottomrule
    \end{tabular}
    \label{tab:app_ablation_geometry_crossview}
\end{table}

\paragraph{Physical parameter sensitivity.}
We conduct a controlled quantitative sensitivity study by fixing all non-physical factors and varying one physical parameter at a time. We report a motion deformation response metric, \emph{Peak Deformation Proxy}, defined as the increase in post-contact spatial spread relative to the pre-contact baseline. For each parameter, we evaluate 30 controlled samples, sweep over 5 parameter values, and repeat each setting 3 times. Table~\ref{tab:app_sensitivity_youngs} and Table~\ref{tab:app_sensitivity_density} report the aggregated normalized deformation response. For Young's modulus, the response decreases as stiffness increases, consistent with the expected trend that stiffer materials deform less under comparable impacts. For density, the response increases monotonically with the density factor, indicating that the generated motion changes consistently with the specified physical parameter.

\begin{table}[H]
    \centering
    \caption{\textbf{Sensitivity analysis for Young's modulus}.}
    \setlength{\tabcolsep}{8pt}
    \renewcommand{\arraystretch}{1.0}
    \begin{tabular}{@{} lccccc @{}}
    \toprule
        Young's Modulus & 1e5 & 3e5 & 1e6 & 3e6 & 1e7 \\ \midrule
        Peak Deformation (norm.) & 1.000 & 0.992 & 0.985 & 0.981 & 0.976 \\
        \bottomrule
    \end{tabular}
    \label{tab:app_sensitivity_youngs}
\end{table}

\begin{table}[H]
    \centering
    \caption{\textbf{Sensitivity analysis for density}.}
    \setlength{\tabcolsep}{8pt}
    \renewcommand{\arraystretch}{1.0}
    \begin{tabular}{@{} lccccc @{}}
    \toprule
        Density Factor & 0.25x & 0.5x & 1.0x & 2.0x & 4.0x \\ \midrule
        Peak Deformation (norm.) & 1.000 & 1.006 & 1.011 & 1.019 & 1.026 \\
        \bottomrule
    \end{tabular}
    \label{tab:app_sensitivity_density}
\end{table}

\paragraph{Dependency on LLM-estimated attributes.}
GPT-4o serves as a practical initializer for physical attributes from the input image and prompt, rather than an infallible source of supervision. To assess the dependency on LLM-estimated attributes, we compare physical attributes estimated by GPT-4o with those provided by human experts. As shown in Table~\ref{tab:app_ablation_expert_attributes}, expert-provided attributes yield only marginal improvement over GPT-4o estimates, suggesting that GPT-4o provides sufficiently reliable physical initialization for OrthoPhys in most cases.

\begin{table}[H]
    \centering
    \caption{\textbf{Ablation on GPT-4o-estimated versus expert-provided physical attributes}.}
    \setlength{\tabcolsep}{8pt}
    \renewcommand{\arraystretch}{1.0}
    \begin{tabular}{@{} lccc @{}}
    \toprule
        Method & Motion↑ & Image↑ & Physical↑ \\ \midrule
        Ours (w/ GPT-4o) & 0.996 & 0.655 & 0.691 \\
        \textbf{Ours (w/ expert)} & \textbf{0.996} & \textbf{0.658} & \textbf{0.694} \\
        \bottomrule
    \end{tabular}
    \label{tab:app_ablation_expert_attributes}
\end{table}

\subsection{Subjective Evaluation with GPT-4o and Human Study.}\label{sec:app_subjective_eval}
\begin{figure}[t]
    \centering
    \includegraphics[width=1\linewidth]{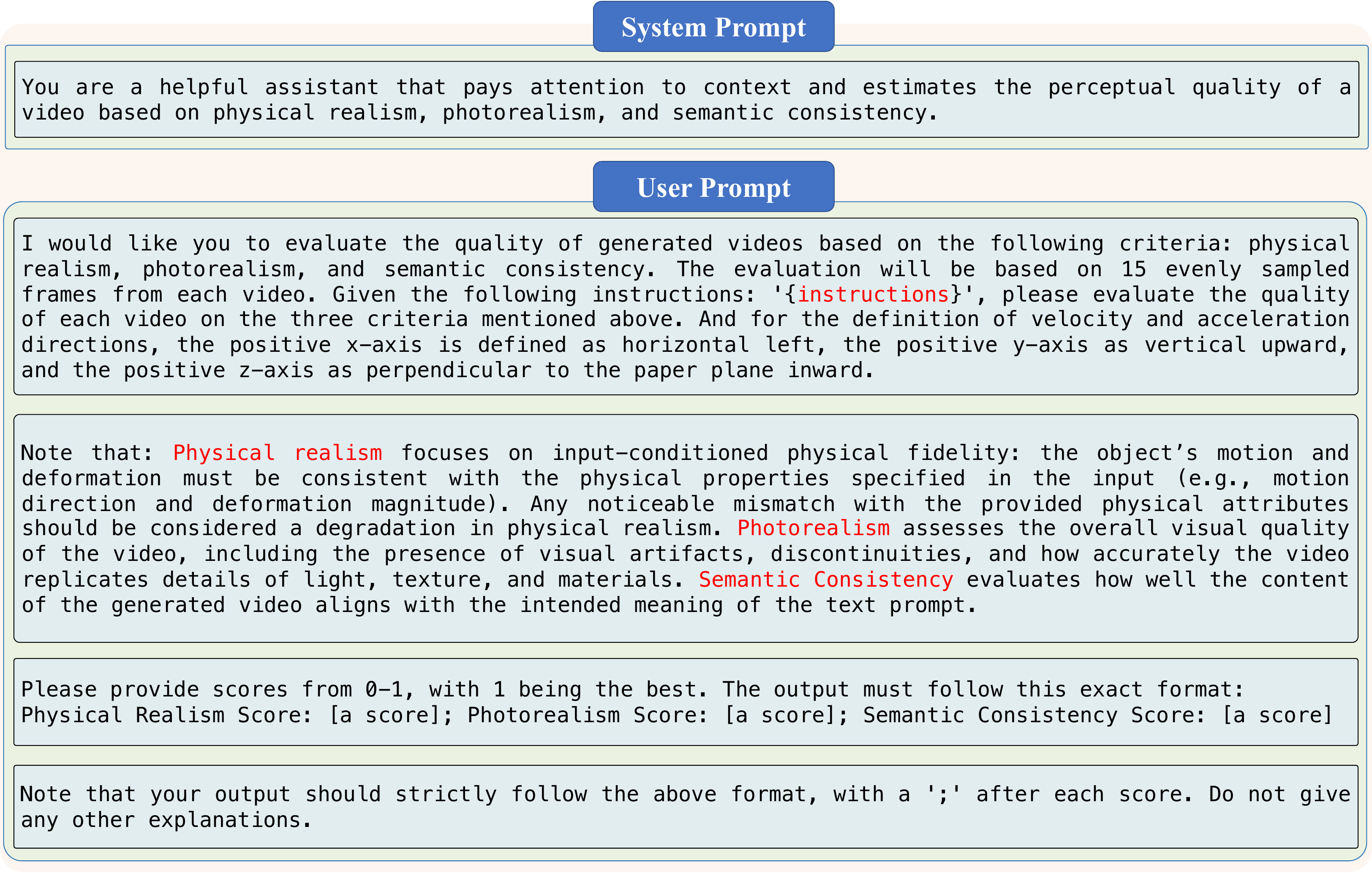}
    \caption{\textbf{Prompt used for GPT-4o evaluation.}}
    \label{fig:prompt}
\end{figure}

Since no widely accepted, physics-focused evaluation metrics for image-to-video generation currently exist, we follow the approach of PhysGen3D~\cite{chen2025physgen3d} and employ GPT-4o to conduct subjective assessments. These evaluations score multiple dimensions relevant to physical plausibility, including physical realism, photorealism, and semantic alignment. We adopt the evaluation prompt design originally proposed in PhysGen3D~\cite{chen2025physgen3d}; as their work has empirically demonstrated strong alignment with human judgment, we apply only minor adaptations to tailor the prompt to our dataset, without introducing substantial modifications. The complete evaluation prompt is illustrated in Fig.~\ref{fig:prompt}. In addition, we conduct a fair blind user study with 59 participants under the same criteria, whose aggregate results are reported in Table~\ref{tab:subjective}. Participants are shown anonymized generated videos in randomized order and asked to score physical realism, photorealism, and semantic consistency on a five-point scale; no personal or sensitive information is collected.

\section{Application of 4D Synthesis}
\begin{figure}[t]
    \centering
    \includegraphics[width=1\linewidth]{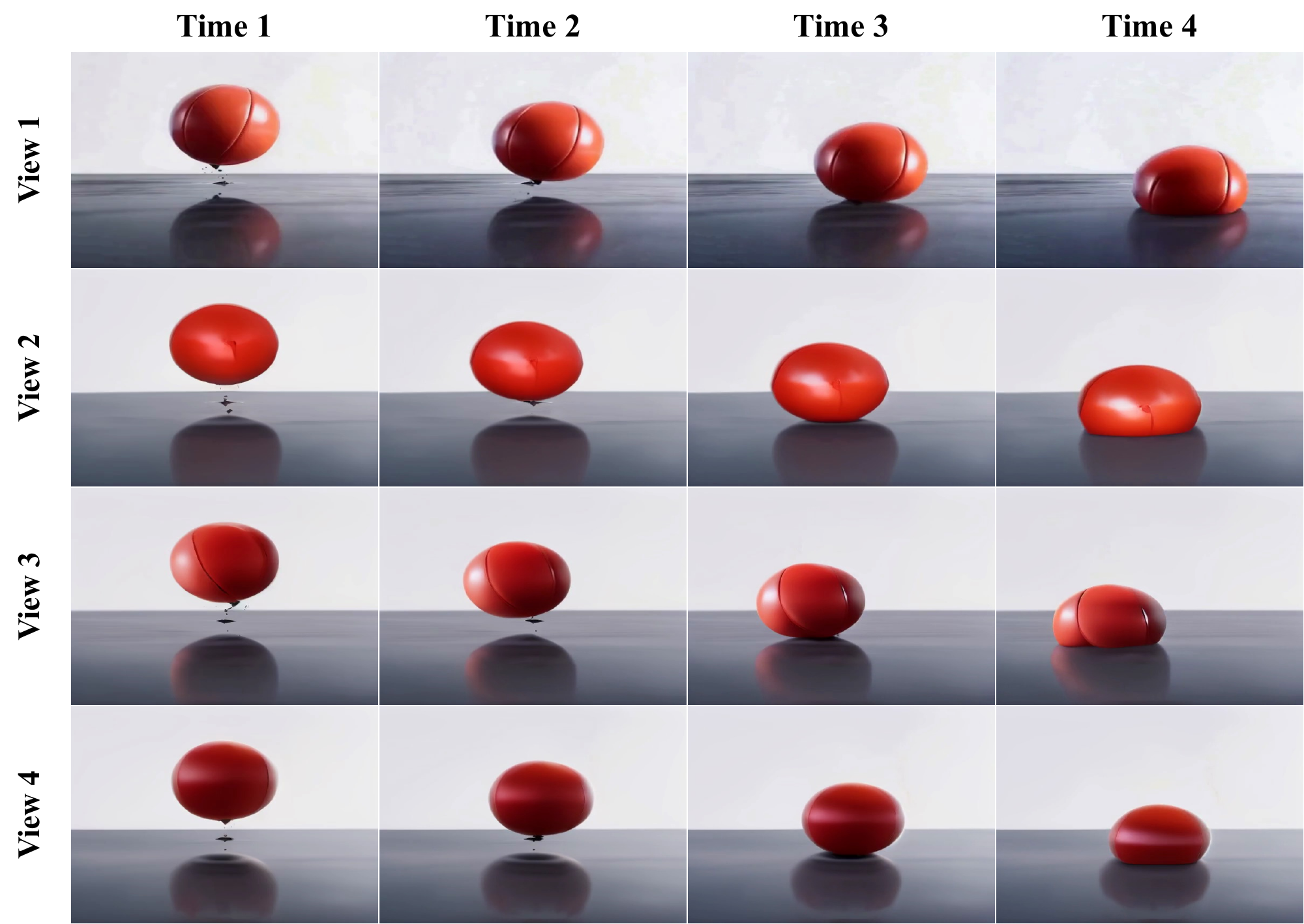}
    \caption{\textbf{Four-view video generation results}. Views 1 through 4 correspond to the front, left, rear, and right viewpoints, respectively.}
    \label{fig:result_4view_bg}
\end{figure}

Our method not only supports object editing while preserving a stable background, but also enables four-view video generation with simple backgrounds. As shown in Fig.~\ref{fig:result_4view_bg}, we render four-view videos from the front, left, back, and right viewpoints. The generated results exhibit strong cross-view consistency, and the interactions between foreground objects and background remain visually plausible. More 4D synthesis results can be viewed on our \href{https://anonymous.4open.science/w/Phys4D/}{project page}.

\section{Details of PhysMV Dataset}\label{sec:app_data}
\begin{figure*}[t]
    \centering
    \includegraphics[width=1\linewidth]{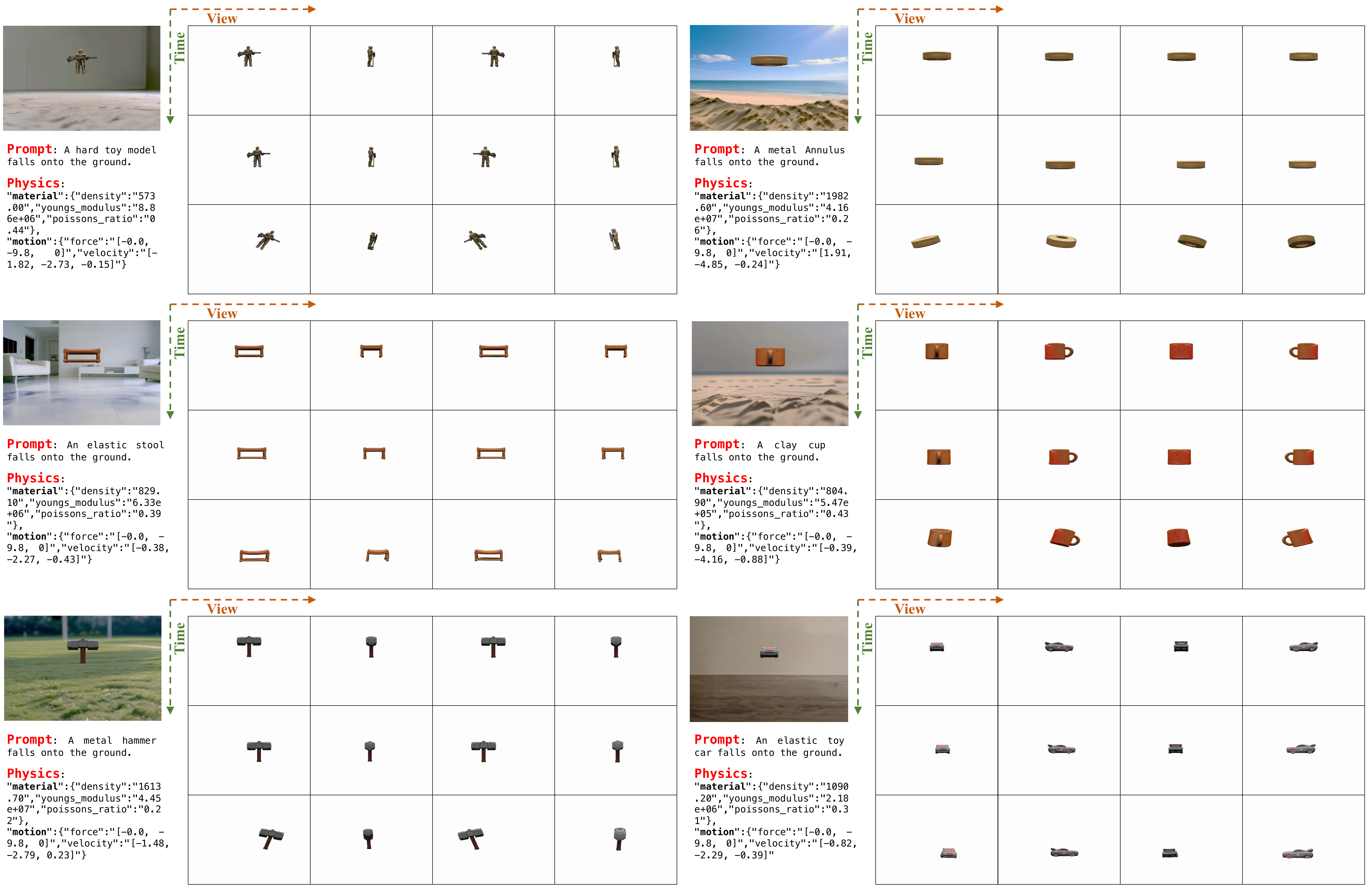}
    \caption{\textbf{Dataset demonstration}. Our dataset contains various objects and motion patterns.}
    \label{fig:supp_dataset}
\end{figure*}

In the main manuscript, we introduce the PhysMV dataset. Here we provide further detailed information. 
We generate 10K 3D objects and reconstruct 10 diverse background scenes, each represented by a 3D-GS model that supports viewpoint exploration. Following the design of OmniPhysGS~\cite{lin2025omniphysgs}, we employ TaiChi~\cite{hu2019taichi} as the physics engine and adopt the Material Point Method (MPM) as the physics solver to simulate object dynamics conditioned on the specified physical properties.
Leveraging the rendering capability of 3D-GS, we render four-view foreground videos after the physical simulation. Note that the generated four-view videos contain only foreground objects, as the background scenes do not support full 360-degree rendering across all views. The dataset examples are shown in Fig.~\ref{fig:supp_dataset}.

For each simulated sample, the textual prompt, motion attributes, and material parameters used to control the physics simulation are automatically recorded as metadata. During post-processing, we apply Grounded-SAM~\cite{ren2024grounded} to obtain semantic foreground masks and DepthAnything V2~\cite{yang2024depth} to get relative depth maps. Finally, Trellis~\cite{xiang2025structured} is again utilized to generate multi-view videos of the segmented foreground objects, which are used to provide multi-view video priors.

\section{More Visual Comparisons with Baselines}\label{More visual comparisons with baselines}
\begin{figure*}[t]
    \centering
    \includegraphics[width=1\linewidth]{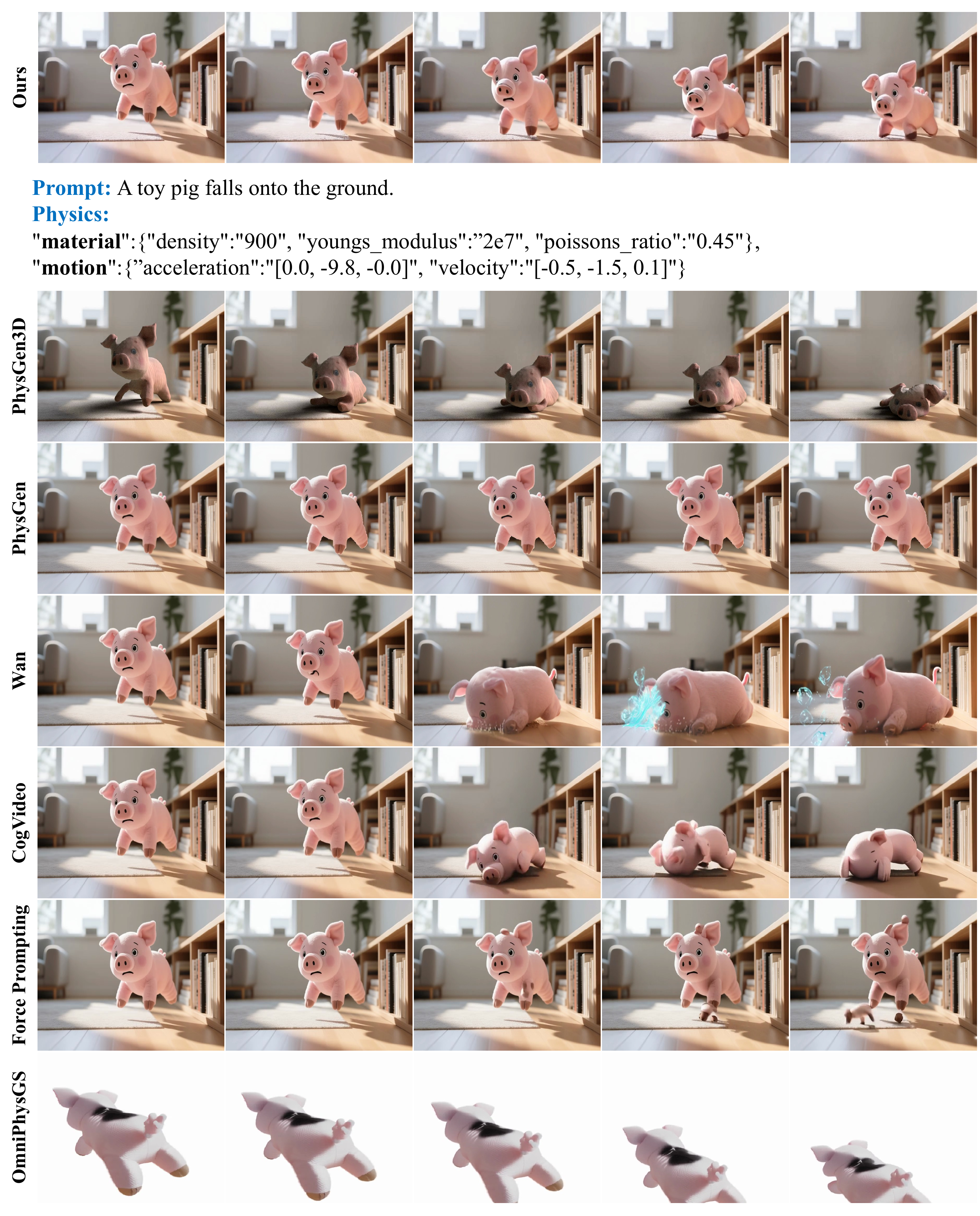}
     \caption{\textbf{Qualitative visualization results of our method and the baselines}.}
    \label{fig:supp_2}
\end{figure*}

\begin{figure*}[t]
    \centering
    \includegraphics[width=1\linewidth]{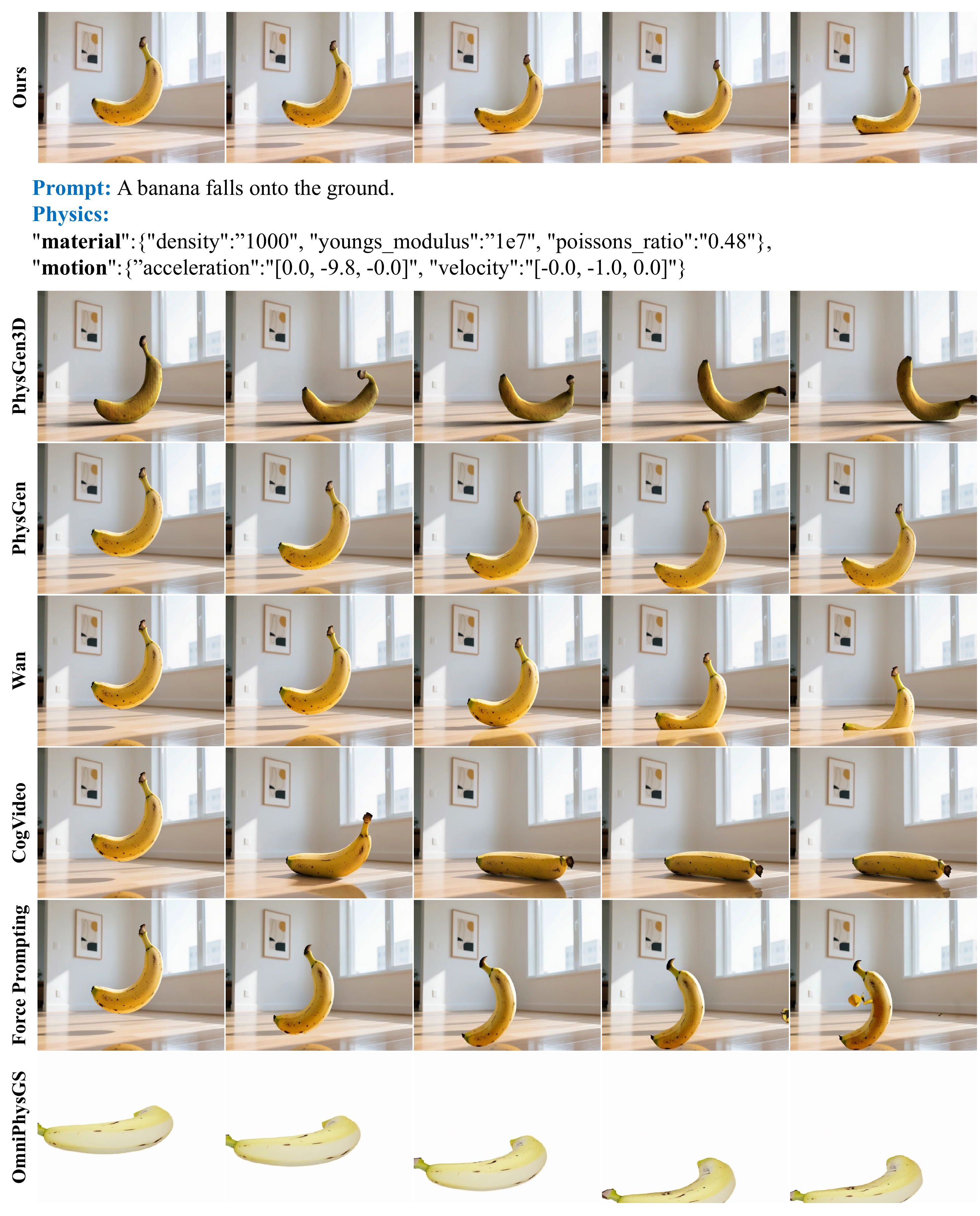}
     \caption{\textbf{Qualitative visualization results of our method and the baselines}.}
    \label{fig:supp_3}
\end{figure*}

\begin{figure*}[t]
    \centering
    \includegraphics[width=1\linewidth]{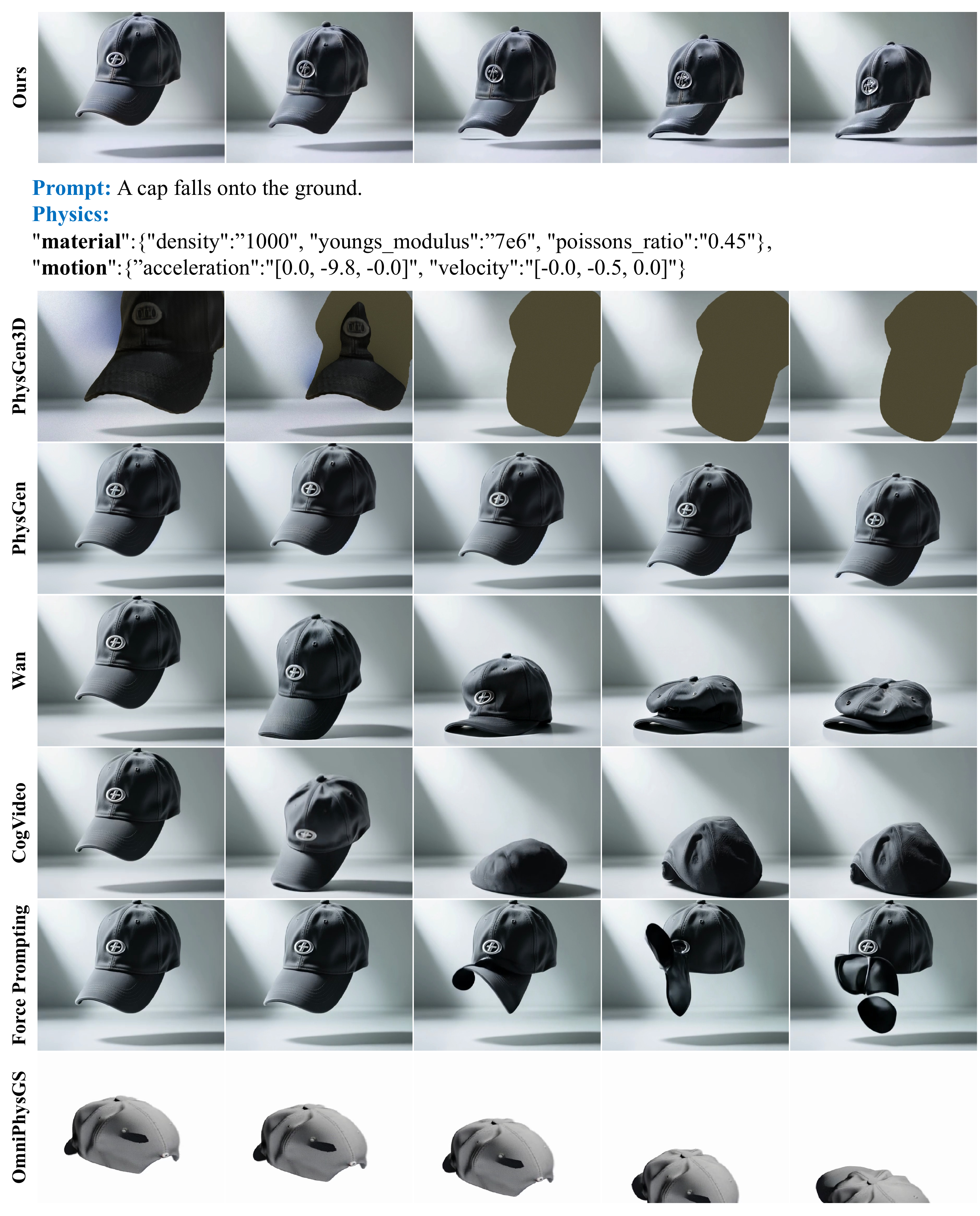}
    \caption{\textbf{Qualitative visualization results of our method and the baselines}.}
    \label{fig:supp_4}
\end{figure*}

\begin{figure*}[t]
    \centering
    \includegraphics[width=1\linewidth]{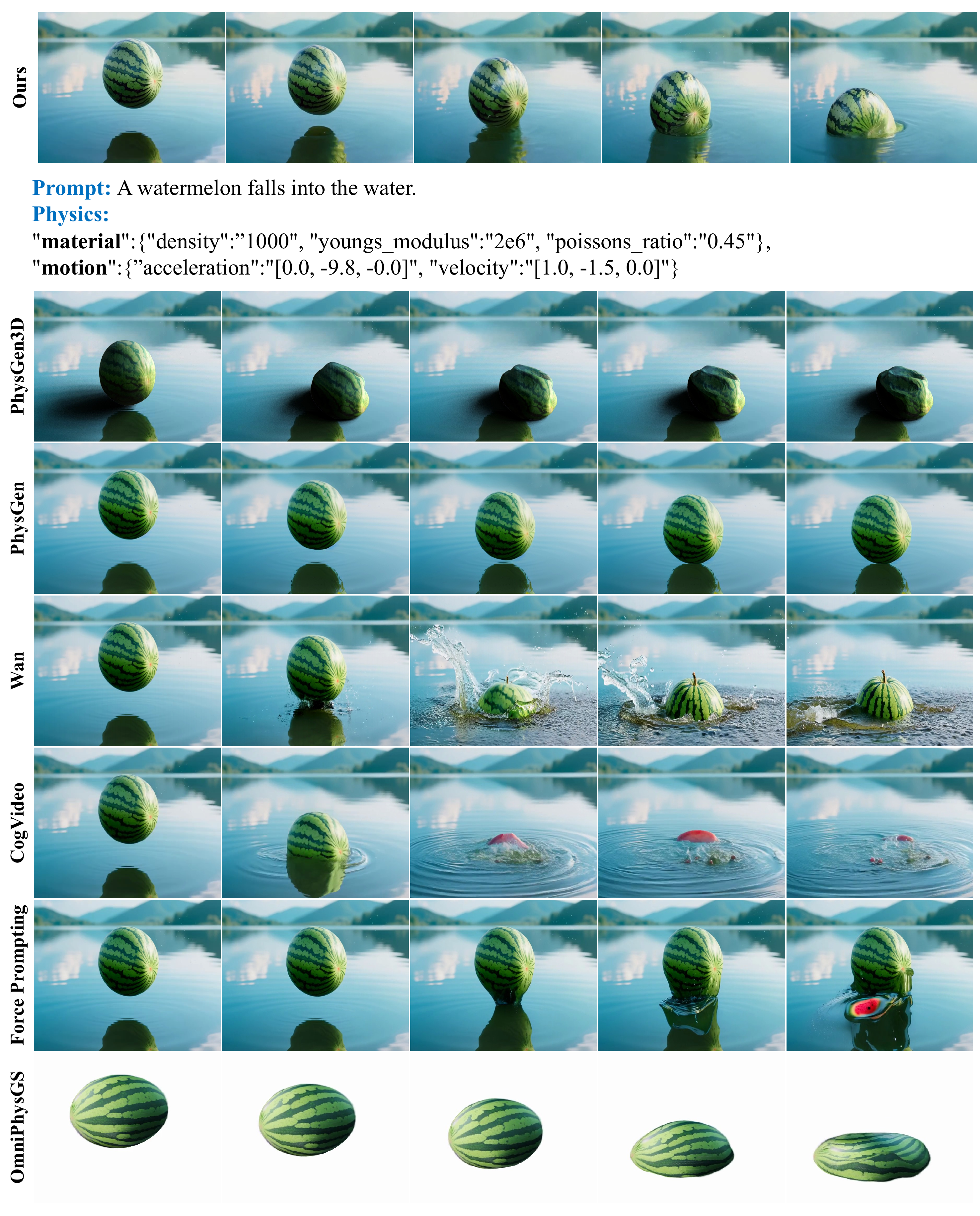}
    \caption{\textbf{Qualitative visualization results of our method and the baselines}.}
    \label{fig:supp_5}
\end{figure*}

As shown in Fig.~\ref{fig:supp_2}, Fig.~\ref{fig:supp_3}, Fig.~\ref{fig:supp_4}, and Fig.~\ref{fig:supp_5}, we provide additional visual comparisons with all baseline methods. These results highlight that our method achieves state-of-the-art quality, characterized by high visual fidelity, strong physical plausibility, and fine-grained controllability.

\section{Limitations and Future Work}\label{sec:limit}
While OrthoPhys introduces a novel framework for generating physics-aware orthogonal-view foreground videos with background interaction from a given viewpoint, it still exhibits several limitations. In particular, our approach is currently constrained in its ability to model and synthesize complex and highly dynamic backgrounds. Although the proposed method can generate diverse and physically plausible orthogonal-view foreground motions, synthesizing complete videos from novel viewpoints remains challenging when the background contains rich geometric structures or intricate appearance variations.
This limitation primarily stems from the inherent difficulty of holistic scene modeling, which has not yet been fully addressed by existing 3D generation and reconstruction methods, and becomes even more challenging in the 4D setting where temporal dynamics must be consistently preserved. As a result, background reconstruction quality may degrade when extrapolating to unseen viewpoints, which in turn affects the overall visual coherence of the generated video.

In future work, we plan to explore more advanced scene representations and background modeling strategies to better capture complex geometry and long-range temporal consistency. Integrating explicit scene decomposition, multi-view supervision, or hybrid 3D-4D representations may help alleviate these limitations and enable more robust whole-scene generation from novel viewpoints.


\end{document}